\documentclass[11pt]{article}

\usepackage{graphics}
\usepackage{graphicx}

\usepackage{amssymb}
\usepackage{amsthm,amsmath}

\usepackage{lineno}

\usepackage{natbib}
\usepackage{url}
\usepackage{multirow}
\usepackage{color}
\usepackage[font=footnotesize]{subcaption}
\usepackage{subcaption}

\usepackage{authblk}
\providecommand{\keywords}[1]{\textbf{\textit{Index terms---}} #1}
\usepackage[a4paper, margin=1.25in]{geometry}
\usepackage{fancyhdr}

\title{Gland Segmentation in Colon Histology Images: The GlaS Challenge Contest}
 
 \author[1]{Korsuk Sirinukunwattana\thanks{k.sirinukunwattana@warwick.ac.uk}}
 
 \author[2]{Josien P. W. Pluim}
 	   
 \author[3]{Hao Chen}
 \author[3]{Xiaojuan Qi}
 \author[3]{Pheng-Ann Heng}

 \author[4]{Yun Bo Guo}
 \author[4]{Li Yang Wang}
 \author[4]{Bogdan J. Matuszewski}
 
 \author[5]{Elia Bruni}
 \author[5]{Urko Sanchez}	
 
 \author[6]{Anton B{\"o}hm} 
 \author[6,7]{Olaf Ronneberger}
 
 \author[8]{Bassem Ben Cheikh}
 \author[8]{Daniel Racoceanu}
 
 \author[9,10]{Philipp Kainz}
 \author[10]{Michael Pfeiffer}
 \author[11,12]{Martin Urschler}

 \author[13]{David R. J. Snead}
 \author[1]{Nasir M. Rajpoot\thanks{n.m.rajpoot@warwick.ac.uk}\thanks{corresponding author}} 
 
 \affil[1]{Department of Computer Science, University of Warwick, Coventry, UK, CV4 7AL}
 
 \affil[2]{Department of Biomedical Engineering, Eindhoven University of Technology, Eindhoven, Netherlands}
 
 \affil[3]{Department of Computer Science and Engineering, The Chinese University of Hong Kong.}
 \affil[4]{School of Engineering, University of Central Lancashire, Preston, UK}
 \affil[5]{ExB Research and Development}
 \affil[6]{Computer Science Department, University of Freiburg, Germany}
 \affil[7]{BIOSS Centre for Biological Signalling Studies, University of Freiburg, Germany and Google-DeepMind, London, UK}
 \affil[8]{Sorbonne Universit\'es, UPMC Univ Paris 06, CNRS, INSERM, Biomedical Imaging Laboratory (LIB), Paris, France}
 \affil[9]{Institute of Biophysics, Center for Physiological Medicine, Medical University of Graz, Graz, Austria}	
 \affil[10]{Institute of Neuroinformatics, University of Zurich and ETH Zurich, Zurich, Switzerland}
 \affil[11]{Institute for Computer Graphics and Vision, BioTechMed, Graz University of Technology, Graz, Austria}
 \affil[12]{Ludwig Boltzmann Institute for Clinical Forensic Imaging, Graz, Austria}
 
 \affil[13]{Department of Pathology, University Hospitals Coventry and Warwickshire, Walsgrave, Coventry, CV2 2DX, UK} 	
\date{}

\begin{document}
\maketitle
\thispagestyle{fancy}
\newpage
\begin{abstract}
Colorectal adenocarcinoma originating in intestinal glandular structures is the most common form of colon cancer. In clinical practice, the morphology of intestinal glands, including architectural appearance and glandular formation, is used by pathologists to inform prognosis and plan the treatment of individual patients. However, achieving good inter-observer as well as intra-observer reproducibility of cancer grading is still a major challenge in modern pathology. An automated approach which quantifies the morphology of glands is a solution to the problem.

This paper provides an overview to the Gland Segmentation in Colon Histology Images Challenge Contest (GlaS) held at MICCAI'2015. Details of the challenge, including organization, dataset and evaluation criteria, are presented, along with the method descriptions and evaluation results from the top performing methods.
\end{abstract}

\keywords{Histology Image Analysis, Segmentation, Colon Cancer, Intestinal Gland, Digital Pathology}



\section{Introduction}

Cancer grading is the process of determining the extent of malignancy and is one of the primary criteria used in clinical practice to inform prognosis and plan the treatment of individual patients. However, achieving good reproducibility in grading most cancers remains one of the challenges in pathology practice \citep{cross2000levels, komuta2004interobserver, fanshawe2008assessing}. With digitized images of histology slides becoming increasingly ubiquitous, digital pathology offers a viable solution to this problem \citep{may2010better}. Analysis of histology images enables extraction of quantitative morphological features, which can be used for computer-assisted grading of cancer making the grading process more objective and reproducible than it currently is \citep{gurcan2009histopathological}. This has led to the recent surge in development of algorithms for histology image analysis.

In colorectal cancer, morphology of intestinal glands including architectural appearance and gland formation is a key criterion for cancer grading \citep{compton2000updated,bosman2010classification,washington2009protocol}. Glands are important histological structures that are present in most organ systems as the main mechanism for secreting proteins and carbohydrates. An intestinal gland (colonic crypt) found in the epithelial layer of the colon, is made up of a single sheet of columnar epithelium, forming a finger-like tubular structure that extends from the inner surface of the colon into the underlying connective tissue \citep{rubin2008rubin,humphries2008colonic}. There are millions of glands in the human colon. Intestinal glands are responsible for absorption of water and nutrients, secretion of mucus to protect the epithelium from a hostile chemical and mechanical environment \citep{gibson1996protective}, as well as being a niche for epithelial cells to regenerate \citep{shanmugathasan2000apoptosis,humphries2008colonic}. Due to the hostile environment, the epithelial layer is continuously regenerating and is one of the fastest regenerating surface in human body \citep{crosnier2006organizing,barker2014adult}. This renewal process requires coordination between cell proliferation, differentiation, and apoptosis. The loss of integrity in the epithelial cell regeneration, through a mechanism that is not yet clearly understood, results in colorectal adenocarcinoma, the most common type of colon cancer.

Manual segmentation of glands is a laborious process. Automated gland segmentation will allow extraction of quantitative features associated with gland morphology from digitized images of CRC tissue slides. Good quality gland segmentation will pave the way for computer-assisted grading of CRC and increase the reproducibility of cancer grading. However, consistent good quality gland segmentation for all the differentiation grades of cancer has remained a challenge. This was a main reason for organizing this challenge contest.

The Gland Segmentation in Colon Histology Images (GlaS) challenge\footnote{\url{http://www.warwick.ac.uk/bialab/GlaScontest}} brought together computer vision and medical image computing researchers to solve the problem of gland segmentation in digitized images of Hematoxylin and Eosin (H\&E) stained tissue slides. Participants developed gland segmentation algorithms, which were applied to benign tissue and to colonic carcinomas. A training dataset was provided, together with ground truth annotations by an expert pathologist. The participants developed and optimized their algorithms on this dataset. The results were judged on the performance of the algorithms on test datasets. Success was measured by how closely the automated segmentation matched the pathologist's.

\section{Related Work}

Recent papers \citep{wu2005b,wu2005a,gunduz2010automatic,Fu,sirinukunwattana,cohen2015memory} indicate the increasing interest in histology image analysis applied to intestinal gland segmentation. In this section, we review some of these methods. 

\citet{wu2005b} presented a region growing method, which first thresholds an image, in order to separate nuclei from other tissue components. Large empty regions, which potentially correspond to lumen found in the middle of glands, are then used to initialize the seed points for region growing. The expanding process for each seed is terminated when a surrounding chain of epithelial nuclei is reached, and subsequently false regions are removed. Although this method performs well in segmenting healthy and benign glands, it is less applicable to cancer cases, where the morphology of glands can be substantially deformed.

In contrast to the above method, which mainly uses pixel-level information, \citet{gunduz2010automatic} represented each tissue component as a disk. Each disk is represented by a vertex of a graph, with nearby disks joined by an edge between the corresponding vertices. They proposed an algorithm, using graph connectivity to identify initial seeds for region growing. To avoid an excessive expansion beyond the glandular region, caused, for example, by large gaps in the surrounding epithelial boundary, edges between nuclear objects are used as a barrier to halt region growing. Those regions that do not show glandular characteristics are eliminated at the last step. The validation of this method was limited only to the dataset with healthy and benign cases.

\citet{Fu} introduced a segmentation algorithm based on polar coordinates. A neighborhood of each gland and a center chosen inside the gland were considered. Using this center to define polar coordinates, the neighborhood  is displayed in $(r,\theta)$ coordinates with the $r$-axis horizontal and the $\theta$-axis vertical. One obtains a vertical strip, periodic with period $2\pi$ in the vertical direction. As a result, the closed glandular boundary is transformed into an approximately vertical periodic path, allowing fast inference of the boundary through a conditional random field model. Support vector regression is later deployed to verify whether the estimated boundary corresponds to the true boundary. The algorithm performs well in both benign and malignant cases stained by Hematoxylin and DAB. However, the validation on routine H\&E stained images was limited only to healthy cases.

\citet{sirinukunwattana} recently formulated a segmentation approach based on Bayesian inference, which allows prior knowledge of the spatial connectivity and the arrangement of neighboring nuclei on the epithelial boundary to be taken into account. This approach treats each glandular structure as a polygon made of a random number of vertices. The idea is based on the observation that a glandular boundary is formed from closely arranged epithelial nuclei. Connecting edges between these epithelial nuclei gives a polygon that encapsulates the glandular structure. Inference of the polygon is made via Reversible-Jump Markov Chain Monte Carlo. The approach shows favorable segmentation results across all histologic grades (except for the undifferentiated grade) of colorectal cancers in H\&E stained images. This method is slow but effective.

Most of the works for intestinal gland segmentation have used different datasets and/or criteria to assess their algorithms, making it difficult to objectively compare their performance. There have been many previous initiatives that provided common datasets and evaluation measures to validate algorithms on various medical imaging modalities \citep{murphy2011evaluation,gurcan2010pattern,roux2013,veta2015assessment}. This not only allows a meaningful comparison of different algorithms but also allows the algorithms to be implemented and configured thoroughly to obtain optimal performance \citep{murphy2011evaluation}. Following these successful initiatives, we therefore organized the Gland Segmentation in Colon Histology Images (GlaS) challenge. This challenge was a first attempt to address the issues of reproducibility and comparability of the results of intestinal gland segmentation algorithms. It was also aimed at speeding up even further the development of algorithms for gland segmentation. Note that none of above methods for intestinal gland segmentation participated in this competition.
\section{Materials} \label{sec:Materials}

\begin{table}[]
\renewcommand{\arraystretch}{1.2}
\renewcommand\tabcolsep{3pt}
\centering
\caption{Details of the dataset.}
\label{tab:breakdownDetials}
\resizebox{0.8\columnwidth}{!}{
\begin{tabular}{|c|c|c|c|}
\hline
\multirow{2}{*}{Histologic Grade} & \multicolumn{3}{c|}{Number of Images (Width x Height in Pixels)} \\ \cline{2-4} 
                                  & Training Part   & Test Part A   & Test Part B   \\ \hline
Benign                            &    $37\; \begin{cases} \;1&(574\times433)\\\;1&(589\times453)\\ \;35&(775\times522)\end{cases}$             &  $33\;\begin{cases} \;1&(574\times433) \\ \;4&(589\times453)\\ \;28&(775\times522) \end{cases}$             & $4\;(775\times522)$   \\ \hline
Malignant                        &  $48\;\begin{cases} \;1&(567\times430)\\\;3&(589\times453)\\\;44&(775\times522)\end{cases}$                &   $27\; \begin{cases} \;1&(578\times433)\\\;2&(581\times442)\\\;24&(775\times522) \end{cases}$            & $16\;(775\times522)$  \\ \hline
\end{tabular}}
\end{table}
\begin{figure}
\centering
\begin{subfigure}{\textwidth}
\includegraphics[width=0.49\linewidth]{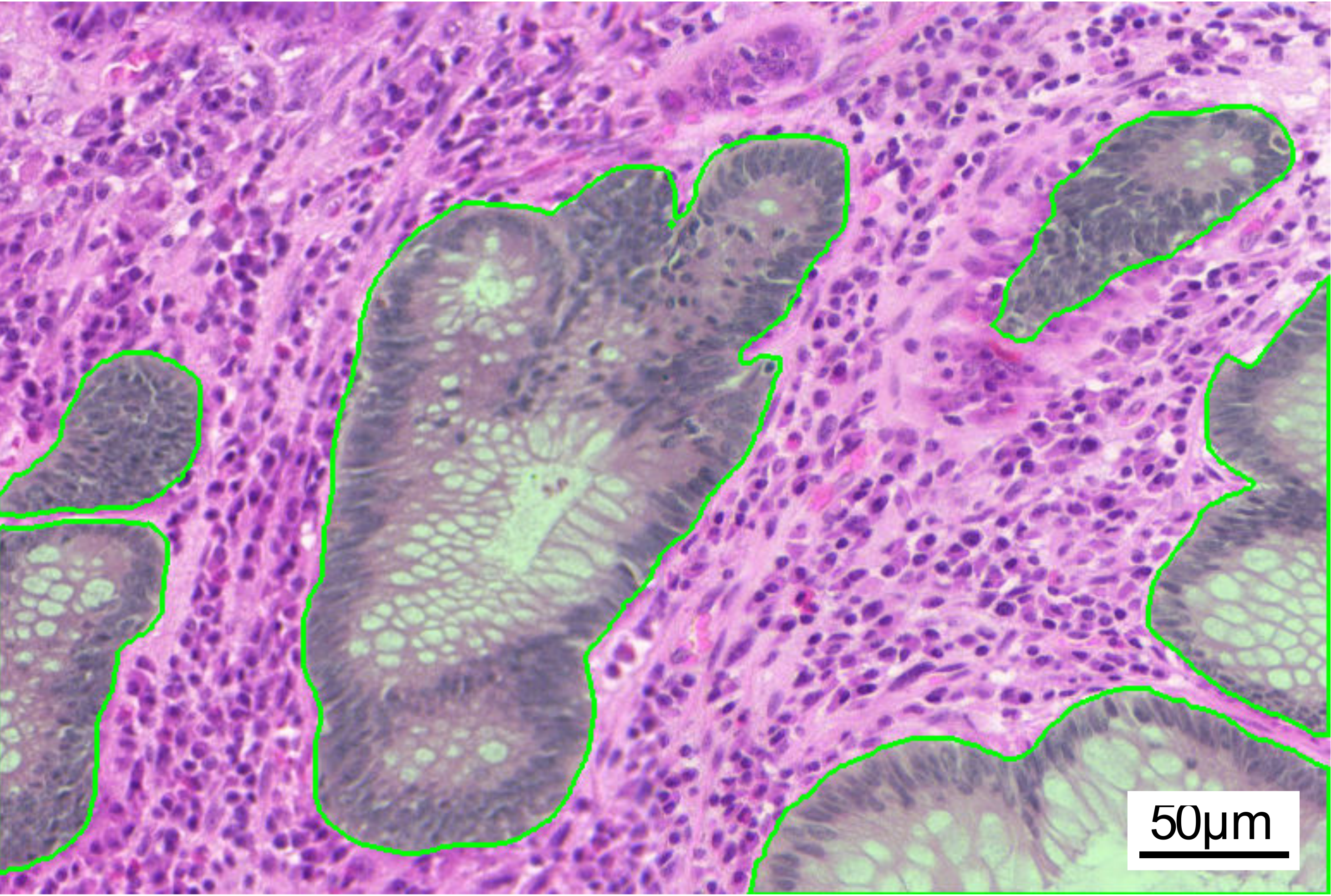}
\hspace*{\fill}
\includegraphics[width=0.49\linewidth]{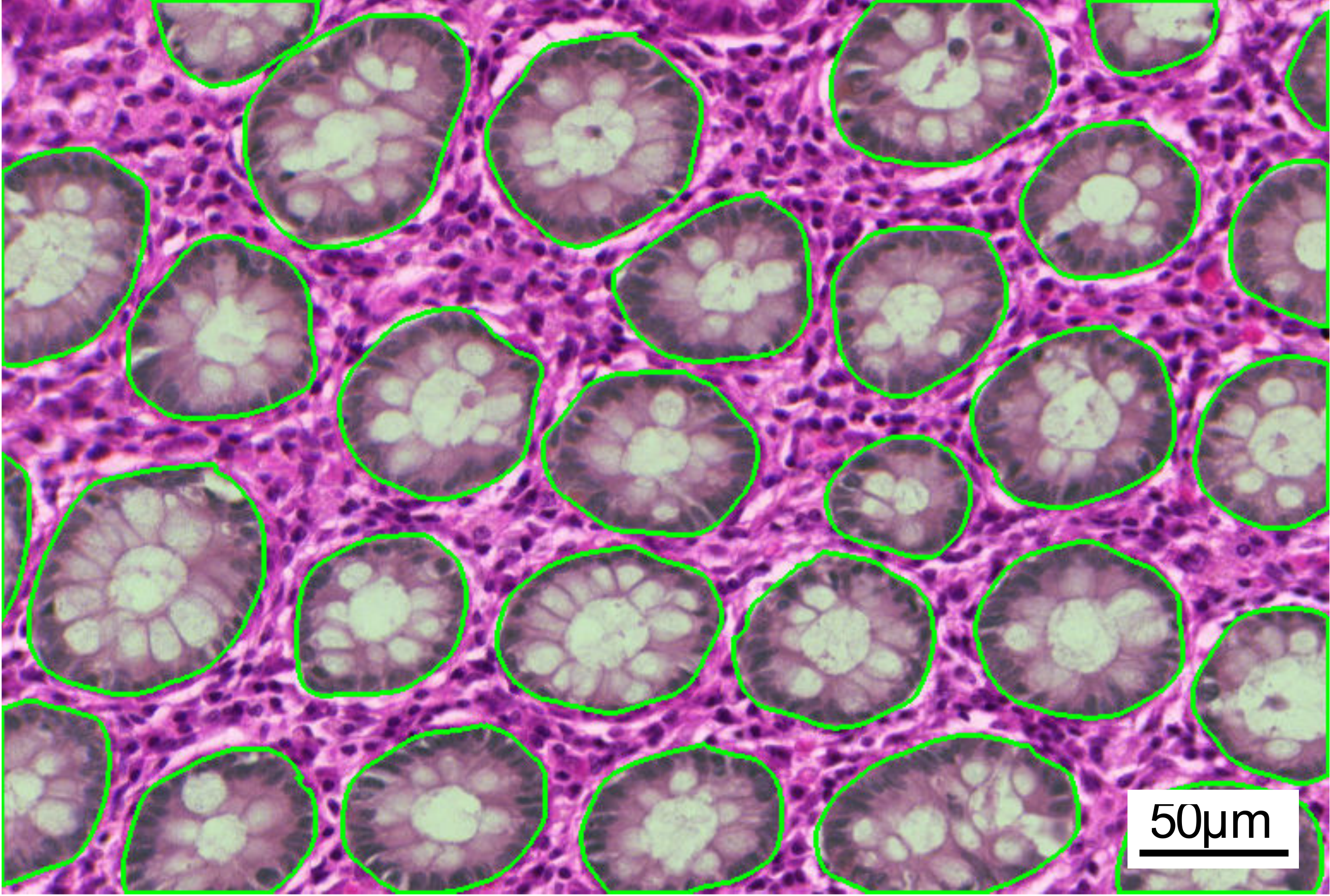}
\caption{}
\label{fig:benign}
\end{subfigure}\\
\begin{subfigure}{\textwidth}
\includegraphics[width=0.49\linewidth]{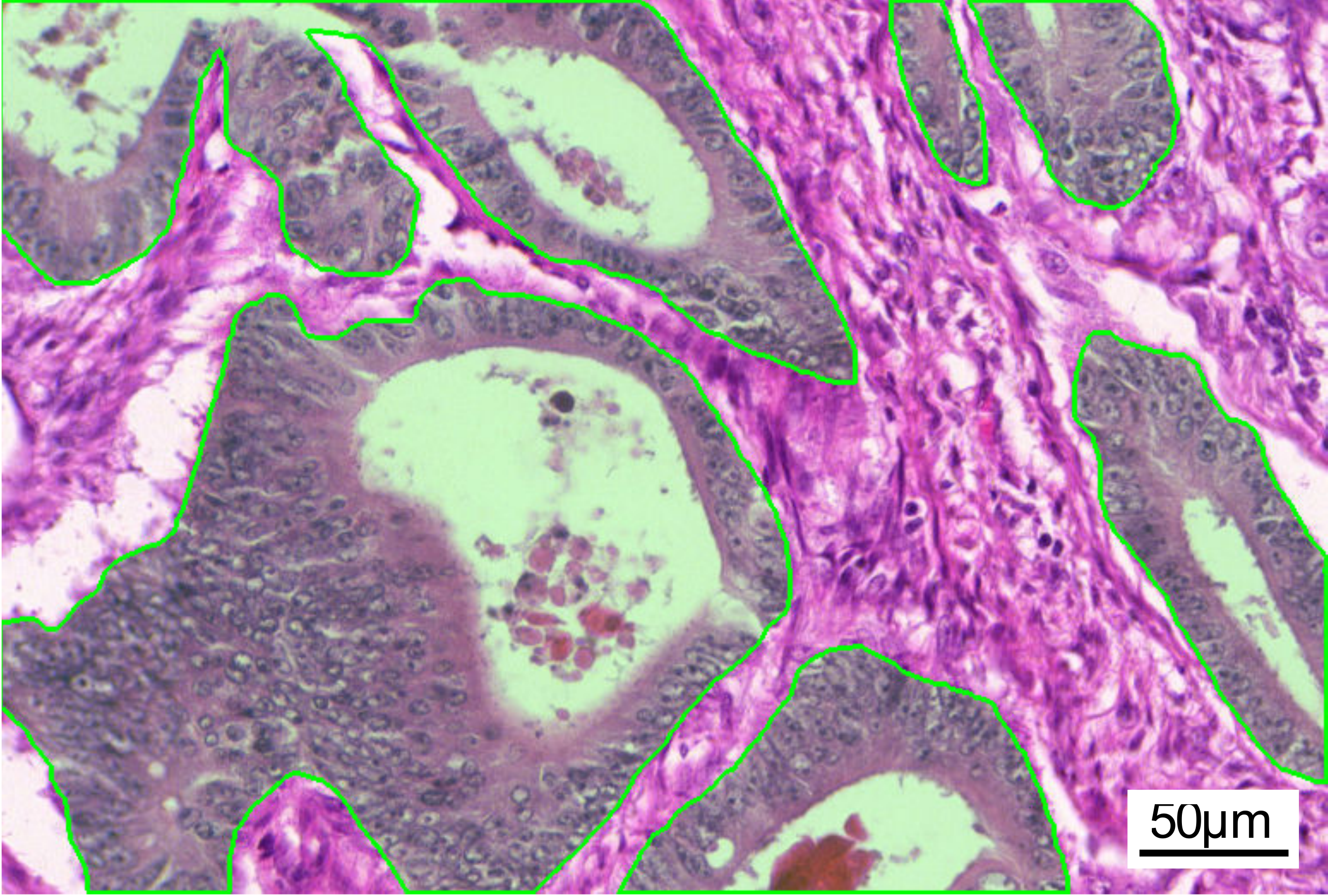}
\hspace*{\fill}
\includegraphics[width=0.49\linewidth]{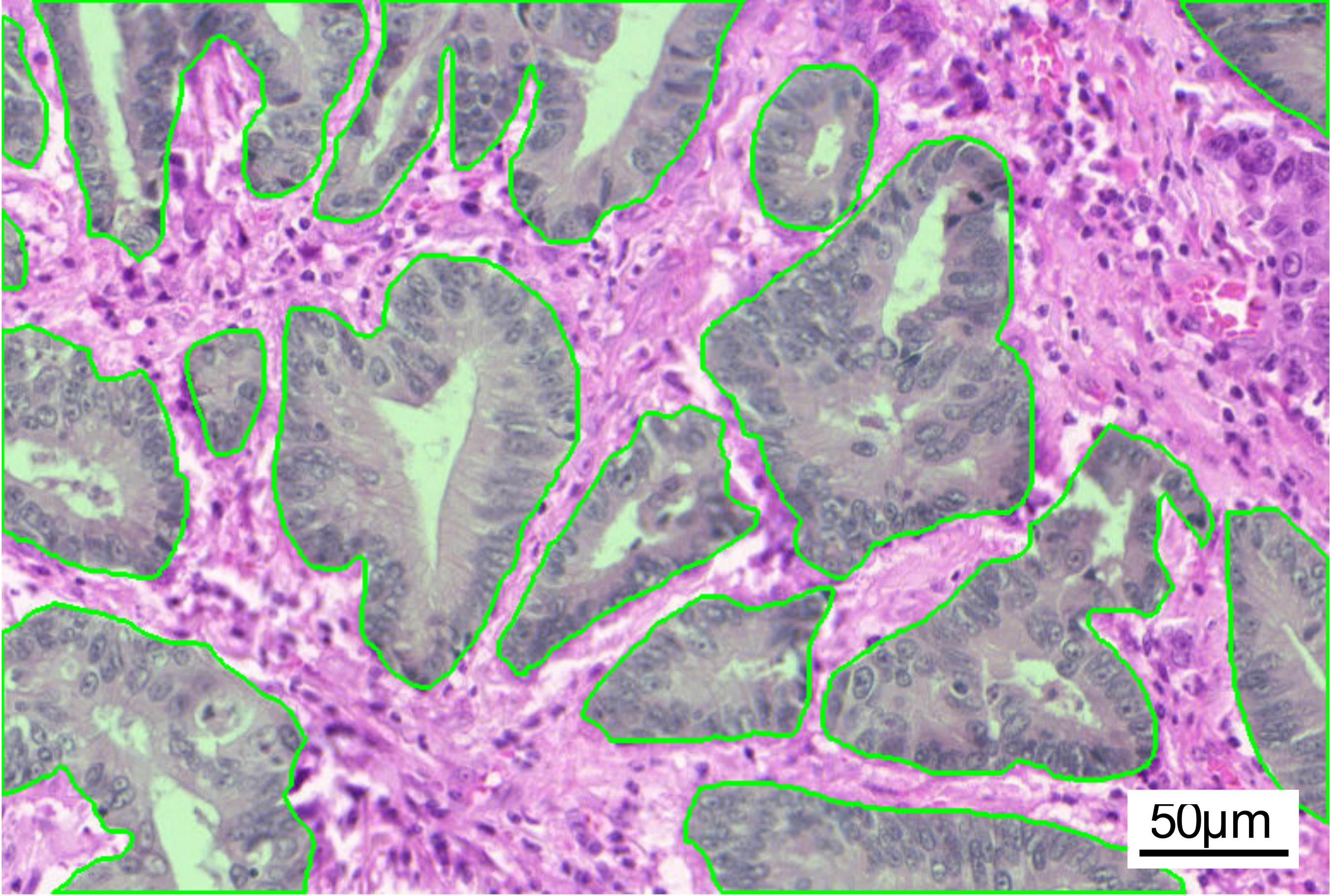}
\caption{}
\label{fig:malignant}
\end{subfigure} 
\caption{Example images of different histologic grades in the dataset: (a) benign and (b) malignant.} \label{fig:example}
\end{figure} 
The dataset used in this challenge consists of 165 images derived from 16 H\&E stained histological sections of stage T3 or T4\footnote{The T in TNM cancer staging refers to the spread of the primary tumour. In colorectal cancer, stage T3 means the tumour has grown into the outer lining of the bowel wall, whereas stage T4 means the tumour has grown through the outer lining of the bowel wall. The cancer stage is different from the tumour histologic grade, as the latter indicates the aggressiveness of the tumour.} colorectal adenocarcinoma. Each section belongs to a different patient, and sections were processed in the laboratory on different occasions. Thus, the dataset exhibits high inter-subject variability in both stain distribution and tissue architecture. The digitization of these histological sections into whole-slide images (WSIs) was accomplished using a Zeiss MIRAX MIDI Slide Scanner with a pixel resolution of $0.465 \mu m$. The WSIs were subsequently rescaled to a pixel resolution of $0.620 \mu m$ (equivalent to $20\times$ objective magnification). 

A total of 52 visual fields from both malignant and benign areas across the entire set of the WSIs were selected in order to cover as wide a variety of tissue architectures as possible. An expert pathologist (DRJS) then graded each visual field as either `benign' or `malignant', according to the overall glandular architecture. The pathologist also delineated the boundary of each individual glandular object on that visual field. We used this manual annotation as ground truth for automatic segmentation. Note that different glandular objects in an image may be part of the same gland. This is because a gland is a 3-dimensional structure that can appear as separated objects on a single tissue section. The visual fields were further separated into smaller, non-overlapping images, whose histologic grades (i.e. benign or malignant) were assigned the same value as the larger visual field. Representative example images of the two grades can be seen in Figure \ref{fig:example}. This dataset was also previously used in the gland segmentation study by \citet{sirinukunwattana}.

In the challenge, the dataset was separated into {\bf Training Part}, {\bf Test Part A}, and {\bf Test Part B}. Note that the data were stratified according to the histologic grade and the visual field before splitting. This was done to ensure that none of the images from the same visual field appears in different parts of the dataset (i.e. Training, Test Part A, or Test Part B). However, since the data were not stratified based on patient, different visual fields from the same slide can appear in different parts of the dataset. A breakdown of the details of the dataset is shown in Table \ref{tab:breakdownDetials}. The ground truth as well as the histologic grade which reflects morphology of glandular structures were provided for every image in the Training Part at the time of release. We used Test Part A and Test Part B as off-site and on-site test datasets respectively. Furthermore, to ensure blindness of evaluation, the ground truth and histologic grade of each image in the test parts were not released to the participants.

\section{Challenge Organization}
The GlaS challenge contest was officially launched by the co-organizers (KS, JPWP, DRJS, NMR) on April 21$^\text{st}$, 2015, and was widely publicized through several channels. At the same point, a challenge website\footnote{\url{http://www.warwick.ac.uk/bialab/GlaScontest}} was set up to disseminate challenge-related information and to serve as a site for registration, submission of results, and communication between the organizers and contestants. The challenge involved 4 stages, as detailed below:

\paragraph{Stage 1: Registration and Release of the Training Data}
The registration was open for a period of about two months (April 21$^\text{st}$ to June 30$^\text{th}$, 2015). Interested individuals or groups of up to 3 people that were affiliated with an academic institute or an industrial organization could register and download the training data (Training Part, see Section \ref{sec:Materials} for details) to start developing their gland segmentation algorithms. From this point forward, we will refer to a separate individual or a group of registrants as a `team'.

\paragraph{Stage 2: Submission of a Short Paper}
In order to gain access to the first part of the test data, each registered team was required to submit a 2-page document containing a general description of their segmentation algorithms and some preliminary results obtained from running each algorithm on the training data. Each team could submit up to 3 different methods. The intention of this requirement was for the organizers to identify teams who were serious about participating in the challenge. The organizers based their reviews on two criteria: clarity of the method description and soundness of the validation strategy. Segmentation performance was not considered in this review. The submission of this document was due by July 17$^\text{th}$, 2015.

\paragraph{Stage 3: Release of the Test Data Part A and Submission of Segmentation Results} 
The first part of the test data (Test Part A, see Section \ref{sec:Materials} for details) was released on August 14$^\text{th}$, 2015 to those teams selected from the previous stage which also agreed to participate in the GlaS contest. The teams were given a month to further adjust and optimize their segmentation algorithms, and carry out segmentation on Part A of the test data. Each team could hand-in up to 3 sets of results per method submitted in Stage 2. The submission of the segmentation results was due by September 14$^\text{th}$, 2015. Evaluation of the submitted results was not disclosed to the teams until after the challenge event.  

\paragraph{Stage 4: GlaS'2015 Challenge Event}
The event was held in conjunction with MICCAI'2015 on October 5$^\text{th}$, 2015. All teams were asked to produce segmentation results on the second part of the test data (Test Part B, see Section \ref{sec:Materials}) within 45 minutes. The teams could either bring their own machines or conduct an experiment remotely. There was no restriction on the number of machines that the teams could use to produce results. Those teams that could not be present at the event provided implementations of their algorithms with which the organizers carried out the segmentation on their behalf. Each team was also asked to give a short presentation, discussing their work. At the end of the event, the complete evaluation of segmentation results across both parts of the test data was announced, which included a final ranking of the submitted methods. This information is also available on the challenge website.

\subsection{Challenge Statistics}
By the end of Stage 1, a total of 110 teams from different academic and industrial institutes had registered. A total of 21 teams submitted the 2-page document for review in Stage 2, and 20 teams were invited to participate in the GlaS competition event. In Stage 3, only 13 teams submitted results on Part A of the test data in time. Late entries were neither evaluated nor considered in the next stage of the competition. On the day of the challenge event, 11 of the 13 teams that submitted the results on time in Stage 3 attended the on-site competition and presented their work. The organizers carried out the segmentation on behalf of the other two teams that could not be present.
\section{Evaluation}
The performance of each segmentation algorithm was evaluated based on three criteria: 1) accuracy of the detection of individual glands; 2) volume-based accuracy of the segmentation of individual glands; and 3) boundary-based similarity between glands and their corresponding segmentation. It may seem that volume-based segmentation accuracy would entail boundary-based segmentation accuracy between a gland and its segmentation. However, in practice, this is not always the case. The volume-based metric for segmentation accuracy used in this challenge, was defined and calculated using the label that the algorithm had assigned to each pixel, but the boundary-based metric used the position assigned by the algorithm to the boundary of each gland. Pixels labels may be fairly accurate, while the boundary curves are very different. The remainder of this section describes all metrics employed in the evaluation.

We use the concept of a pair of corresponding segmented and ground truth objects as proposed in \citet{sirinukunwattana}. Let $\mathcal{S}$ denote a set of all segmented objects and $\mathcal{G}$ denote a set of all ground truth objects. We also include in each of these sets the empty object $\emptyset$. We define a function $G_*:\mathcal{S}\to \mathcal{G}$, by setting, for each segmented object $S\in \mathcal{S}$, $G_*(S) = G \in \mathcal{G}$ where $G$ has the largest possible overlapping area with $S$. Although there could be more than one $G \in \mathcal{G}$ that maximally overlaps $S$, this in practice is extremely rare, and it is good enough to consider one of these $G$ as the value of $G_*(S)$. If there is no overlapping $G$, we set $G_*(S) = \emptyset$. (However, in the context of Hausdorff distance -- see Section \ref{sec:hausdorff} -- $G_*$ will be extended in a different way.) Similarly, we define $S_*:\mathcal{G}\to\mathcal{S}$, by setting, for each $G\in\mathcal{G}$, $S_*(G)=S\in\mathcal{S}$, where $S$ has the largest possible overlapping area with $G$. Note that $G_*$ and $S_*$ are, in general, neither injective, nor surjective. Nor are they inverse to each other, in general. They do, however, assign to each $G$ an $S=S_*(G)$, and to each $S$ a $G=G_*(S)$.

\subsection{Detection Accuracy}

The F1 score is employed to measure the detection accuracy of individual glandular objects. A segmented glandular object that intersects with at least 50\% of its ground truth object is counted as true positive, otherwise it is counted as false positive. The number of false negatives is calculated as the difference between the number of ground truth objects and the number of true positives. Given these definitions, the F1 score is defined by
\begin{equation} \label{eq:f1score}
\mathrm{F1score} = \frac{2\cdot \mathrm{Precision} \cdot \mathrm{Recall}}{\mathrm{Precision} + \mathrm{Recall}},
\end{equation}
where
\begin{equation}
\mathrm{Precision} = \frac{\mathrm{TP}}{\mathrm{TP} + \mathrm{FP}}, \quad \mathrm{Recall} = \frac{\mathrm{TP}}{\mathrm{TP} + \mathrm{FN}},
\end{equation}
and $\mathrm{TP}, \mathrm{FP}$, and $\mathrm{FN}$ denote respectively the number of true positives, false positives, and false negatives from all images in the dataset.

\subsection{Volume-Based Segmentation Accuracy}

\subsubsection{Object-Level Dice Index} \label{sec:obj-dice}

The Dice index \citep{dice1945measures} is a measure of agreement or similarity between two sets of samples. Given $G$, a set of pixels belonging to a ground truth object, and $S$, a set of pixels belonging to a segmented object, the Dice index is defined as follows:
\begin{equation}\label{eq:dice}
\mathrm{Dice}(G,S) = \frac{2 |G \cap S|}{|G| + |S|},
\end{equation}
where $|\cdot|$ denotes set cardinality. The index ranges over the interval $[0,1]$, where the higher the value, the more concordant the segmentation result and the ground truth. A Dice index of 1 implies a perfect agreement. It is conventional that the segmentation accuracy on an image is calculated by $\mathrm{Dice}(G_\text{all},S_\text{all})$, where $G_\text{all}$ denotes the set of pixels of all ground truth objects and $S_\text{all}$ denotes the set of pixels of all segmented objects. The calculation made in this way measures the segmentation accuracy only at the pixel level, not at the gland level, which was the main focus of the competition.

To take the notion of an individual gland into account, we employ the object-level Dice index \citep{sirinukunwattana}. Let $n_\mathcal{G}$ be the number of non-empty ground truth glands, as annotated by the expert pathologist.  Similarly let $n_\mathcal{S}$ be the number of glands segmented by the algorithm, that is the number of non-empty segmented objects. Let $G_i \in \mathcal{G}$ denote the $i^\text{th}$ ground truth object, and let $S_j \in \mathcal{S}$ denote the $j^\text{th}$ segmented object. The object-level Dice index is defined as
\begin{equation}\label{eq:obj-dice}
\mathrm{Dice}_\mathrm{obj}(\mathcal{G},\mathcal{S}) = \frac{1}{2}\left[\sum_{i= 1}^{n_\mathcal{G}}\gamma_i \mathrm{Dice}(G_i,S_*(G_i)) + \sum_{j=1}^{n_\mathcal{S}} {\sigma}_j \mathrm{Dice}(G_*(S_j),S_j)\right], 
\end{equation}
where 
\begin{equation}
\gamma_i = |G_i| / \sum_{p=1}^{n_\mathcal{G}} |G_p|,  \quad {{\sigma}}_j = |S_j|/\sum_{q = 1}^{n_\mathcal{S}}|S_q| 
\end{equation}
On the right hand side of \eqref{eq:obj-dice}, the first summation term reflects how well each ground truth object overlaps its segmented object, and the second summation term reflects how well each segmented object overlaps its ground truth objects. Each term is weighted by the relative area of the object, giving less emphasis to small segmented and small ground truth objects.

In the competition, the object-level Dice index of the whole test dataset was calculated by including all the ground truth objects from all images in $\mathcal{G}$ and all the segmented objects from all images in $\mathcal{S}$.

\subsubsection{Adjusted Rand Index}
We also included the adjusted Rand index \citep{hubert1985comparing} as another evaluation measure of segmentation accuracy. This index was used for additional assessment of the algorithm performance in Section \ref{sec:AdditionalExperiment}.

The adjusted Rand index measures similarity between the set of all ground truth objects $\mathcal{G}$ and the set of all segmented objects $\mathcal{S}$, based on how pixels in a pair are labeled. Two possible scenarios  for the pair to be concordant are that (i) they are placed in the same ground truth object in $\mathcal{G}$ and the same segmented object in $\mathcal{S}$, and (ii) they are placed in different ground truth objects in $\mathcal{G}$ and in different segmented objects in $\mathcal{S}$. Define $n_{ij}$ as the number of pixels that are common to both the $i^\text{th}$ ground truth object and the $j^\text{th}$ segmented object, $n_{i,\cdot}$ as the total number of pixels in the $i^\text{th}$ ground truth object,  $n_{\cdot,j}$ as the total number of pixels in the $j^\text{th}$ segmented object, and $n$ as the total number of pixels. Following a simple manipulation, it can be shown that the probability of agreement is equal to
\begin{equation}
P_\text{agreement} = \left[\binom{n}{2} + 2\sum_{i=1}^{n_\mathcal{G}}\sum_{j=1}^{n_\mathcal{S}} \binom{n_{ij}}{2} - \sum_{i = 1}^{n_\mathcal{G}}\binom{n_{i,\cdot}}{2} - \sum_{j=1}^{n_\mathcal{S}}\binom{n_{\cdot,j}}{2}  \right]\Bigg/\binom{n}{2}.
\end{equation}
Here, the numerator term corresponds to the total number of agreements, while the denominator term corresponds to the total number of all possible pairs of pixels. Under the assumption that the partition of pixels into ground truth objects in $\mathcal{G}$ and segmented objects in $\mathcal{S}$   follows a generalized hypergeometric distribution, the adjusted Rand index can be formulated as
\begin{equation}
\mathrm{ARI}(\mathcal{G},\mathcal{S}) = \frac{\sum_{i=1}^{n_\mathcal{G}} \sum_{j=1}^{n_\mathcal{S}}\binom{n_{i,j}}{2} - \sum_{i=1}^{n_\mathcal{G}}\binom{n_i}{2}\sum_{j=1}^{n_\mathcal{S}}\binom{n_{\cdot,j}}{2}\big/\binom{n}{2}}{\frac{1}{2}\left[\sum_{i=1}^{n_\mathcal{G}} \binom{n_{i,\cdot}}{2} + \sum_{j=1}^{n_\mathcal{S}} \binom{n_{\cdot,j}}{2} \right] - \sum_{i=1}^{n_\mathcal{G}} \binom{n_{i,\cdot}}{2}\sum_{j=1}^{n_\mathcal{S}}\binom{n_{\cdot,j}}{2} \big/ \binom{n}{2} }.
\end{equation}
The adjusted Rand index is bounded above by 1, and it can be negative.

\subsection{Boundary-Based Segmentation Accuracy} \label{sec:hausdorff}
We measure the boundary-based segmentation accuracy between the segmented objects in $\mathcal{S}$ and the ground truth objects in $\mathcal{G}$ using the object-level Hausdorff distance. The usual definition of a Hausdorff distance between ground truth object $G$ and segmented object $S$ is 
\begin{equation} \label{eq:hausdorff}
\mathrm{H}(G,S) = \max\{\sup_{x\in G} \inf_{y \in S} d(x, y), \sup_{y\in S} \inf_{x\in G} d(x,y)\}
\end{equation}
where $d(x,y)$ denotes the distance between pixels $x\in G$ and $y\in S$. In this work, we use the Euclidean distance. According to \eqref{eq:hausdorff}, Hausdorff distance is the most extreme value from all distances between the pairs of nearest pixels on the boundaries of $S$ and $G$. Thus, the smaller the value of the Hausdorff distance, the higher the similarity between the boundaries of $S$ and $G$, and $S=G$ if their Hausdorff distance is zero.

To calculate the overall segmentation accuracy between a pair of corresponding segmented and ground truth objects, we now introduce object-level Hausdorff distance by imitating the definition of object-level Dice index \eqref{eq:obj-dice}. The object-level Hausdorff distance is defined as
\begin{equation}
\mathrm{H}_\text{obj}(\mathcal{G},\mathcal{S}) = \frac{1}{2}\left[\sum_{i=1}^{n_\mathcal{G}} \gamma_i \mathrm{H}(G_i,S_*(G_i)) + \sum_{j=1}^{n_\mathcal{S}}{\sigma}_j \mathrm{H}(G_*(S_j),S_j)\right],
\end{equation}
where the meaning of the mathematical notation is similar to that given in Section \ref{sec:obj-dice}. In case a ground truth object $G$ does not have a corresponding segmented object (i.e. $S_*(G) = \emptyset$), the Hausdorff distance is calculated between $G$ and the nearest segmented object $S\in\mathcal{S}$ to $G$ (in the Hausdorff distance) in that image instead. The same applies for a segmented object that does not have a corresponding ground truth object.
\section{Ranking Scheme} \label{sec:RankingScheme}
 
Each submitted entry was assigned one ranking score per evaluation metric and  set of test data. Since there were 3 evaluation metrics (F1 score for gland detection, object-level Dice index for volume-based segmentation accuracy, and object-level Hausdorff index for boundary-based segmentation accuracy) and 2 sets of test data, the total number of ranking scores was 6. The best performing entry was assigned ranking score 1, the second best was assigned ranking score 2, and so on. In care of a tie, the standard competition ranking was applied. For instance, F1 score 0.8, 0.7, 0.7, and 0.6 would result in the ranking scores 1, 2, 2, and 4. The final ranking was then obtained by adding all 6 ranking scores (rank sum). The entry with smallest sum was placed top in the final ranking.
\section{Methods} \label{sec:Methods} 
The top ranking methods are described in this section. They are selected from the total of 13 methods that participated in all stages of the challenge. The cut-off for the inclusion in this section was made where there was a substantial gap in the rank sums (see \ref{sec:appendixA}, Figure \ref{fig:glas2}). Of the 7 selected methods, only 6 preferred to have their methods described here.

\subsection[CUMedVision]{CUMedVision\footnote{Department of Computer Science and Engineering, The Chinese University of Hong Kong.}}
A novel deep contour-aware network \citep{chen2016dcan} was presented. This method explored the multi-level feature representations with fully convolutional networks (FCN) \citep{long2015fully}. The network outputted segmentation probability maps and depicted the contours of gland objects simultaneously. The network architecture consisted of two parts: a down-sampling path and an up-sampling path. The down-sampling path contained convolutional and max-pooling layers while the up-sampling path contained convolutional and up-sampling layers, which increased the resolutions of feature maps and outputted the prediction masks. In total, there were 5 max-pooling layers and 3 up-sampling layers. Each layer with learned parameters was followed by a non-linear mapping layer (element-wise rectified linear activation).

In order to separate touching glands, the feature maps from hierarchical layers were up-sampled with two different branches to output the segmented object and contour masks respectively. The parameters of the down-sampling path were shared and updated for these two kinds of masks. This could be viewed as a multi-task learning framework with feature representations, simultaneously encoding the information of segmented objects and contours. To alleviate the problem of insufficient training data \citep{chen2015automatic}, an off-the-shelf model from DeepLab \citep{chen2014semantic}, trained on the 2012 PASCAL VOC dataset\footnote{\url{http://host.robots.ox.ac.uk:8080/pascal/VOC/voc2012/index.html}}, was used to initialize the weights for layers in the down-sampling path. The parameters of the network were obtained by minimizing the loss function with standard back-propagation \footnote{More details will be available at: \url{http://www.cse.cuhk.edu.hk/~hchen/research/2015miccai_gland.html}}.

The team submitted two entries for evaluation. \textbf{CUMedVision1} was produced by FCN with multi-level feature representations relying only on gland object masks, while \textbf{CUMedVision2} was the results of the deep contour-aware network, which considers gland object and contour masks simultaneously.

\subsection[CVML]{CVML\footnote{School of Engineering, University of Central Lancashire, Preston, UK.}}
In the first, preprocessing, stage the images were corrected to compensate for variations in the appearance due to a variability of the tissue staining process. This was implemented through histogram matching, where the target histogram was calculated from the whole training data, and the individual image histograms were used as inputs. The main processing stage was based on two methods: a convolutional neural network (CNN) \citep{krizhevsky2012imagenet} for a supervised pixel classification, and a level set segmentation for grouping pixels into spatially coherent structures. The employed CNN used an architecture with two convolutional, pooling and fully connected layers. The network was trained with three target classes. The classes were designed to represent (1) the tubular interior of the glandular structure (inner class), (2) epithelial cells forming boundary of the glandular structure (boundary class) and (3) inter-gland tissue (outer class). The inputs to the CNN were $19\times19$ pixel patches sliding across the adjusted RGB input image. The two convolutional layers used $6\times6$ and $4\times4$ kernels with 16 and 36 feature maps respectively. The pooling layers, implementing the mean function, used $2\times2$ receptive fields and $2\times2$ stride. The first and second fully connected layers used the rectified linear unit and softmax functions respectively. The outputs from the CNN were two probability maps representing the probability of each image pixel belonging to the inner and boundary classes. These two probability maps were normalized between -1 and 1 and used as a propagation term, along with an advection term and a curvature flow term. These terms were part of the hybrid level set model described in \citet{zhang2008medical}. In the post-processing stage, a sequence of morphological operations was performed to removed small objects, fill holes and disconnect weakly connected objects. Additionally, if an image boundary intersecting an object forms a hole, the corresponding pixels was labeled as part of that object. The team submitted a single entry for evaluation, henceforth referred to as \textbf{CVML}.

\subsection[ExB]{ExB\footnote{ExB Research and Development.}} 

This method first preprocessed the data by performing per channel zero mean and unit variance normalization, where the mean and variance were computed from the training data. The method then exploited the local invariance properties of the task by applying a set of transformations to the data. At training time, the dataset was augmented by applying affine transformations, Gaussian blur and warping. During testing, both image mirroring and rotation were applied. 

The main segmentation algorithm consisted of a multi-path convolutional neural network. Each path was equipped with a different set of convolutional layers and configured to capture features from different views in a local-global fashion. All the different paths were connected to a set of two fully connected layers. A leaky rectified linear unit was used as a default activation function between layers, and a softmax layer was used after the last fully connected layer.  Every network was trained via stochastic gradient descent with momentum, using a step-wise learning rate schedule \citep{krizhevsky2012imagenet}. The network was randomly initialized such that unit variance was preserved across layers. It was found that using more than three paths led to heavy over-fitting -- this was due to insufficient training data.

Simple-path networks were trained to detect borders of glands. The ground truth for these networks was constructed using a band of width $K \in [5,10]$ pixels along a real gland border. These values of $K$  were found to produce optimal and equivalent quantitative results, measured by the F1 score and the object-Dice index. The output of these networks was used to better calibrate the final prediction.  

In the post-processing step, a simple method was applied to clean noise and fill holes in the structures. Thresholding was applied to remove spurious structures with diameter smaller than a certain epsilon. Filling-hole criteria based on diameter size was also used. 

Using the initial class discrimination (benign and malignant), a simple binary classifier constructed from a convolutional neural network with 2 convolutional and 1 fully connected layers was trained. This binary classifier used the raw image pixels as input. The output of the classifier was used together with the border networks and the post-processing method to apply a different set of parameters/thresholds depending on the predicted class. The hyperparameters for the entire pipeline, including post-processing and border networks, were obtained through cross-validation.  

For this method, the team submitted 3 entries. \textbf{ExB 1} was a two-path network including both the border network for detecting borders of glands and the binary classification to differentiate between the post-processing parameters. \textbf{ExB 2} was similar to ExB 1 without the use of the border network. \textbf{ExB 3} used a two-path network without any post-processing. 

\subsection[Image Analysis Lab Uni Freiburg]{Image Analysis Lab Uni Freiburg\footnote{Computer Science Department and BIOSS Centre for Biological Signalling Studies, University of Freiburg, Germany.}}
The authors applied a u-shaped deep convolutional network ``u-net''\footnote{The implementation of the u-net is freely available at \url{http://lmb.informatik.uni-freiburg.de/people/ronneber/u-net/}.} \citep{RFB} for the segmentation. The input was the raw RGB image and the output was a  binary segmentation map (glands and background). The network consisted of an analysis-path constructed from a sequence of convolutional layers and max-pooling layers, followed by a synthesis path with a sequence of up-convolutional layers and convolutional layers, resulting in 23 layers in total. Additional shortcut-connections propagated the feature maps at all detail levels from the analysis to the synthesis path. The network was trained from scratch in an end-to-end fashion with only the images and ground truth segmentation maps provided by the challenge organizers. To teach the network the desired invariances and to avoid overfitting, the training data were augmented with randomly transformed images and the correspondingly transformed segmentation maps. The applied transformations were random elastic deformations, rotations, shifts, flips, and blurs. The color transformations were random multiplications applied in the HSV color space. To avoid accidentally joining touching objects, a high pixel-wise loss weight was introduced for pixels in thin gaps between objects in the training dataset (see \citet{RFB}). The exact same u-net layout with the same hyperparameters as in \citet{RFB} was used for the challenge. The only difference were more training iterations and a slower decay of the learning rate.  

The team submitted two entries. The first entry \textbf{Freiburg1} was a connected component labelling applied to the raw network output.  The second entry \textbf{Freiburg2} post-processed the segmentation maps with morphological hole-filling and deletion of segments smaller than 1000 pixels. 

\subsection[LIB]{LIB\footnote{Sorbonne Universit\'es, UPMC Univ Paris 06, CNRS, INSERM, Biomedical Imaging Laboratory (LIB), Paris, France.}}
Intestinal glands were divided according to their appearance into three categories: hollow, bounded, and crowded. A hollow gland was composed of lumen and goblet cells and it could be a hole in the tissue surface. A bounded gland had the same composition, but in addition, it was surrounded by a thick epithelial layer. A crowded gland was composed of bunches of epithelial cells clustered together and it might have shown necrotic debris. 

The tissue was first classified into one of the above classes before beginning the segmentation. The classification relied on the characterization of the spatial distribution of cells and the topology of the tissue. Therefore, a closing map was generated with a cumulative sum of morphological closing by a disk of increasing radius (1 to 40 pixels) on the binary image of nuclear objects, which were segmented by the $k$-means algorithm in the RGB colour space. The topological features were calculated from a normalized closing map in MSER fashion (Maximally Stable Extremal Region, \citet{matas2004robust}) as the number of regions below three different thresholds (25\%, 50\% and 62.5\%) and above one threshold (90\%), their sizes and the mean of their corresponding values in the closing map. The first three thresholds characterized the holes and the fourth one characterized the thickness of nuclear objects. After classifying the tissue with a Naive Bayes classifier trained on these features, a specific segmentation algorithm was applied.

Three segmentation algorithms were presented, one for each category.  Hollow glands were delineated by morphological dilation on regions below 50\%. Bounded gland candidates were first detected as hollow glands, then the thickness of nuclear objects surrounding the region was evaluated by generating a girth map and a solidity map \citep{bassem}, then after classifying nuclear objects, the epithelial layer was added or the candidate was removed. Crowded glands were identified as populous regions (regions above 90\%), and then morphological filtering was applied for refinement. The team submitted a single entry labeled as \textbf{LIB} for evaluation.

\subsection[vision4GlaS]{vision4GlaS\footnote{Institute of Biophysics, Center for Physiological Medicine, Medical University of Graz, Graz, Austria; Institute of Neuroinformatics, University of Zurich and ETH Zurich, Zurich, Switzerland; Institute for Computer Graphics and Vision, BioTechMed, Graz University of Technology, Graz, Austria; Ludwig Boltzmann Institute for Clinical Forensic Imaging, Graz, Austria.}}
Given an H\&E-stained RGB histopathological section, the gland segmentation method was based on a pixel-wise classification and an active contour model, and it proceeded in three steps \citep{kainz2015semantic}. In a first preprocessing step the image was rescaled to half the spatial resolution, and color deconvolution separated the stained tissue components. The red channel of the deconvolved RGB image represented the tissue structure best and was therefore considered for further processing. Next, two convolutional neural networks (CNNs) \citep{lecun2010convolutional} of seven layers each were trained for pixel-wise classification on a set of image patches. Each network was trained with ReLU nonlinearities, and stochastic gradient descent with momentum, weight decay, and dropout regularization to minimize a negative log-likelihood loss function. The first CNN, called Object-Net, was trained to distinguish four classes: (i) benign background, (ii) benign gland, (iii) malignant background, and (iv) malignant gland. For each image patch the probability distribution over the class labels was predicted, using a softmax function. The Object-Net consisted of three convolutional layers followed by max-pooling, a final convolutional layer and three fully connected layers. The second -- architecturally similar -- CNN called Separator-Net, learned to predict pixels of gland-separating structures in a binary classification task. Ground truth was generated by manually labeling image locations, close to two or more gland borders, as gland-separating structures. In the final step the segmentation result was obtained by combining the outputs of the two CNNs. Predictions for benign and malignant glands were merged, and predictions of gland-separating structures were subtracted to emphasize the foreground probabilities. Background classes were handled similarly. Using these refined foreground and background maps, a figure-ground segmentation based on weighted total variation was employed to find a globally optimal solution. This approach optimized a geodesic active contour energy, which minimized contour length while adhering to the refined CNN predictions \citep{bresson2007fast}. The team submitted a single entry, referred to as \textbf{vision4GlaS}.
\section{Results and Discussion}

\subsection{Summary of the Methods}
The methods described above take one of the following two approaches to segmentation: (a) they start by identifying pixels corresponding to glands which are then grouped together to form separated, spatially coherent objects; (b) they begin with candidate objects that are then classified as glands or non-glands. All methods that are based on CNNs (CUMedVision, CVML, ExB, Freiburg, and vision4GlaS) follow the former approach. CVML, ExB, and vision4GlaS built CNN classifiers that assign a gland-related or non-gland-related label to every pixel in an image, by taking patch(es) centered at the pixel as input. ExB, in particular, use multi-path networks into which patches at different sizes are fed, in order to capture contextual information at multiple scales. CUMedVision and Freiburg, on the other hand, base their pixel classifier on a fully convolutional network architecture \citep{long2015fully}, allowing simultaneous pixel-wise label assignment at multiple pixel locations. To separate gland-related pixels into individual objects, CVML and vision4GlaS deploy contour based approaches. ExB trains additional networks for glandular boundary, while CUMedVision and Freiburg explicitly include terms for boundary in the training loss function of their networks. The only method that follows the latter approach for object segmentation is LIB. In this method, candidate objects forming part of a gland (i.e., lumen, epithelial boundary) are first identified, and then classified into different types, followed by the final step of segmentation.

A variety of data transformation and augmentation were employed to deal with variation within the data. In order to counter the effect of stain variation, CVML and ExB performed transformations of the RGB color channels, vision4GlaS used a stain deconvolution technique to obtain only the basophilic channel in their preprocessing step. By contrast, Freiburg tackled the issue of stain variability through data augmentation, which implicitly forces the networks to be robust to stain variation to some extent. As is common among methods using CNNs, spatial transformations, such as affine transformations (e.g. translation, rotation, flip), elastic deformations (e.g. pincushion and barrel distortions), and blurring, were also used in the data augmentation to teach the network to learn features that are spatially invariant. The other benefit of data augmentation is it provides, to some extent, avoidance of over-fitting.

ExB, LIB, and vision4GlaS incorporated histologic grades of glands in their segmentation approach. In ExB, procedures and/or parameter values used in boundary detection and post-precessing steps were different, subject to the predicted histologic grade of an image. vision4GlaS classified pixels based on histological information. Although not explicit, LIB categorized candidate objects forming glands according to their appearance, related to histologic grades, before treating them in different ways. 

As a post-processing step, many segmentation algorithms employed simple criteria and/or a sequence of morphological operations to improve their segmentation results. A common treatment was to eliminate small spurious segmented objects. Imperfections in pixel labelling can result in the appearance of one or more holes in the middle of an object. Filling such holes is often necessary. In addition to these operations, CVML performed morphological operations to separate accidentally joined objects.


\subsection{Evaluation Results} \label{sec:EvaluationResults}
\begin{table}[]
\renewcommand{\arraystretch}{1.2}
\renewcommand\tabcolsep{3pt}
\centering
\caption{Summary results. The evaluation is carried out according to the  challenge criteria described in Section \ref{sec:RankingScheme}. A ranking score is assigned to each algorithm according to its performance in each evaluation measure, obtained from each test part. The entries are listed in a descending order based on their rank sum}
\label{tab:completeResults}
\resizebox{\columnwidth}{!}{
\begin{tabular}{|c|c|c|c|c|c|c|c|c|c|c|c|c|c|}
\hline
\multirow{3}{*}{Method} & \multicolumn{4}{c|}{$\mathrm{F1score}$}                             & \multicolumn{4}{c|}{$\mathrm{Dice}_\mathrm{obj}$}                                 & \multicolumn{4}{c|}{$\mathrm{H}_\mathrm{obj}$}                            & \multirow{3}{*}{Rank Sum} \\ \cline{2-13}
                        & \multicolumn{2}{c|}{Part A} & \multicolumn{2}{c|}{Part B} & \multicolumn{2}{c|}{Part A} & \multicolumn{2}{c|}{Part B} & \multicolumn{2}{c|}{Part A} & \multicolumn{2}{c|}{Part B} &                           \\ \cline{2-13}
                        & Score         & Rank        & Score         & Rank        & Score         & Rank        & Score         & Rank        & Score          & Rank       & Score          & Rank       &                           \\ \hline
CUMedVision2            & 0.912         & 1           & 0.716         & 3           & 0.897         & 1           & 0.781         & 5           & 45.418         & 1          & 160.347        & 6          & 17                        \\ \hline
ExB1                    & 0.891         & 4           & 0.703         & 4           & 0.882         & 4           & 0.786         & 2           & 57.413         & 6          & 145.575        & 1          & 21                        \\ \hline
ExB3                    & 0.896         & 2           & 0.719         & 2           & 0.886         & 2           & 0.765         & 6           & 57.350         & 5          & 159.873        & 5          & 22                        \\ \hline
Freiburg2               & 0.870         & 5           & 0.695         & 5           & 0.876         & 5           & 0.786         & 3           & 57.093         & 3          & 148.463        & 3          & 24                        \\ \hline
CUMedVision1            & 0.868         & 6           & 0.769         & 1           & 0.867         & 7           & 0.800         & 1           & 74.596         & 7          & 153.646        & 4          & 26                        \\ \hline
ExB2                    & 0.892         & 3           & 0.686         & 6           & 0.884         & 3           & 0.754         & 7           & 54.785         & 2          & 187.442        & 8          & 29                        \\ \hline
Freiburg1               & 0.834         & 7           & 0.605         & 7           & 0.875         & 6           & 0.783         & 4           & 57.194         & 4          & 146.607        & 2          & 30                        \\ \hline
CVML                    & 0.652         & 9           & 0.541         & 8           & 0.644         & 10          & 0.654         & 8           & 155.433        & 10         & 176.244        & 7          & 52                        \\ \hline
LIB                     & 0.777         & 8           & 0.306         & 10          & 0.781         & 8           & 0.617         & 9           & 112.706        & 9          & 190.447        & 9          & 53                        \\ \hline
vision4GlaS             & 0.635         & 10          & 0.527         & 9           & 0.737         & 9           & 0.610         & 10          & 107.491        & 8          & 210.105        & 10         & 56                        \\ \hline
\end{tabular}}
\end{table}

\begin{figure}
\centering
\renewcommand{\arraystretch}{1.2}
\renewcommand\tabcolsep{3pt}
\begin{tabular}{ccc}
\includegraphics[width=0.32\linewidth]{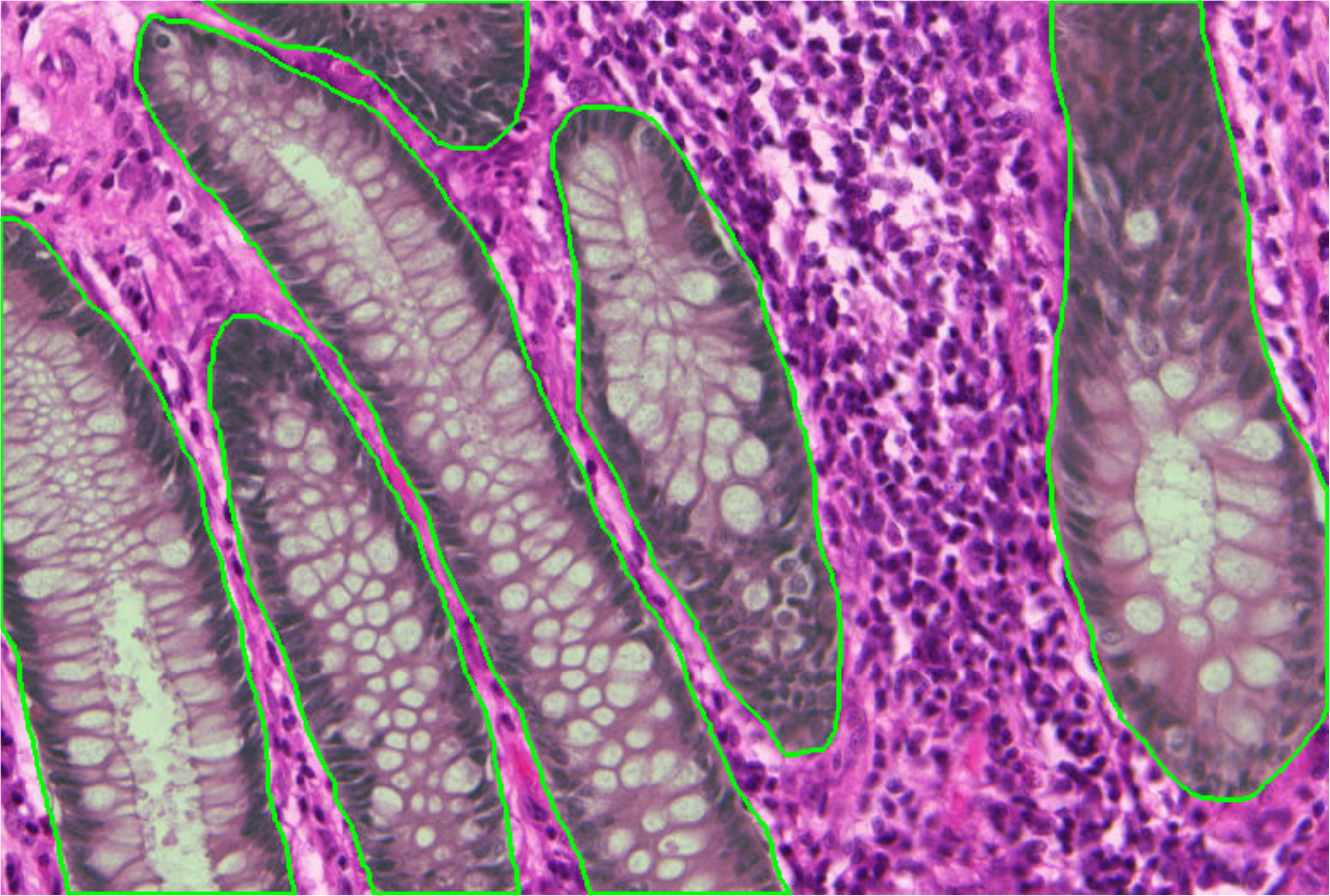} &
\includegraphics[width=0.32\linewidth]{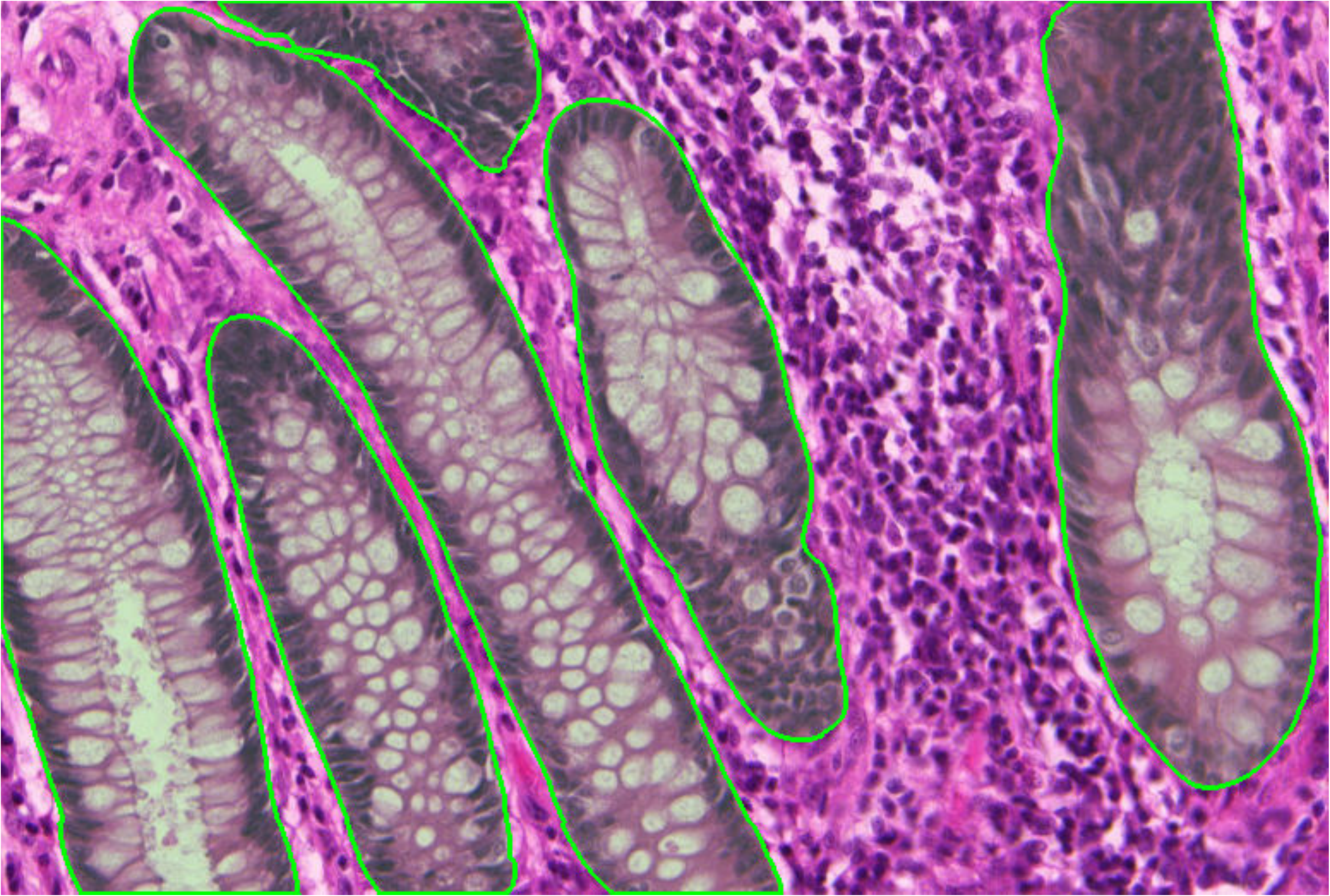} &
\includegraphics[width=0.32\linewidth]{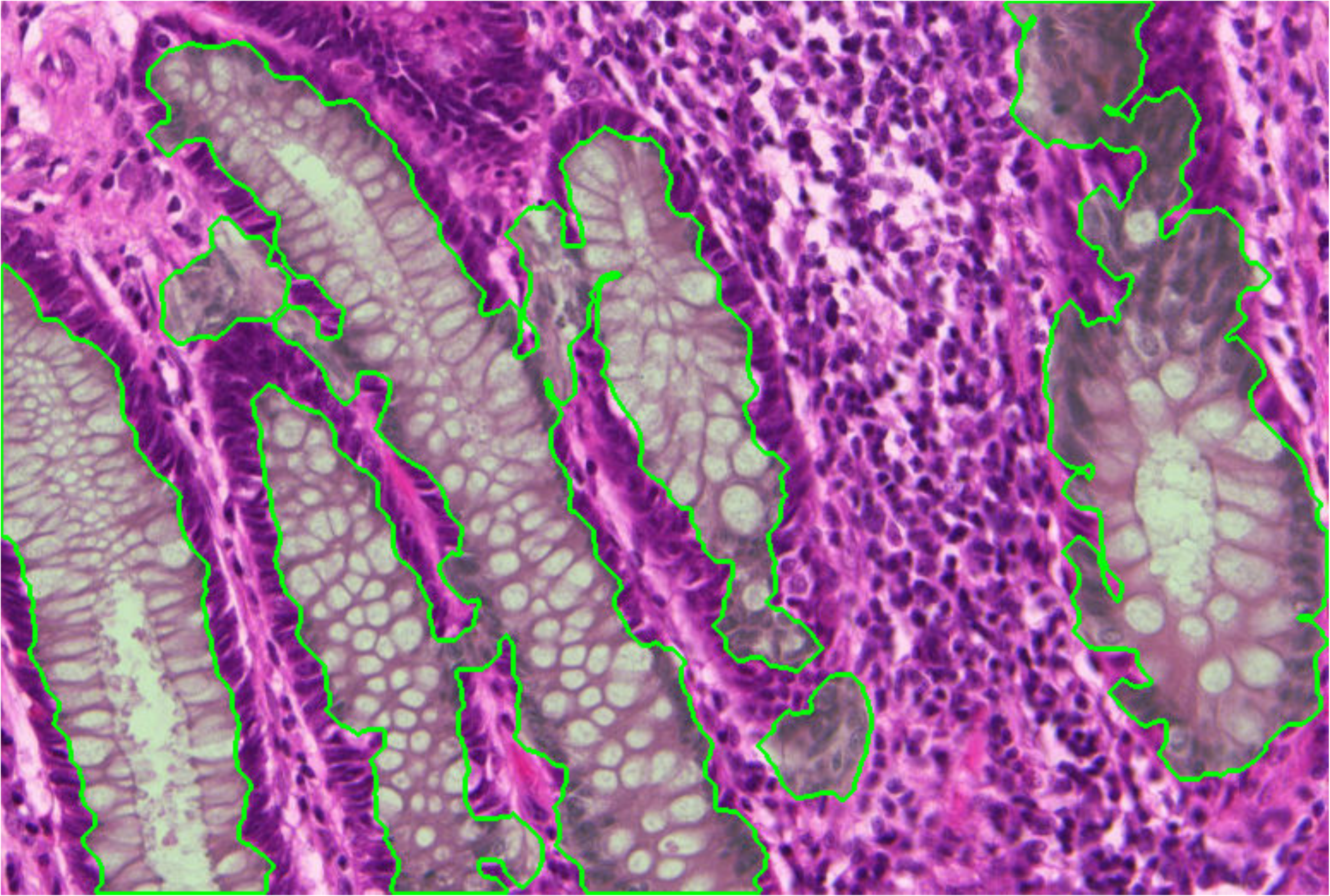}	 \\
& \renewcommand{\arraystretch}{0.8}
\renewcommand\tabcolsep{3pt}
\begin{tabular}{c}
{\footnotesize $\mathrm{F1score} = 1.000$}\\
{\footnotesize $\mathrm{Dice}_\mathrm{obj} = 0.969$}\\
{\footnotesize $\mathrm{H}_\mathrm{obj} = 10.322$}
\end{tabular} & 
\renewcommand{\arraystretch}{0.8}
\renewcommand\tabcolsep{3pt}
\begin{tabular}{c}
{\footnotesize $\mathrm{F1score} = 0.546$} \\
{\footnotesize $\mathrm{Dice}_\mathrm{obj} = 0.661$} \\
{\footnotesize $\mathrm{H}_\mathrm{obj} = 107.580$}
\end{tabular} \\
\includegraphics[width=0.32\linewidth]{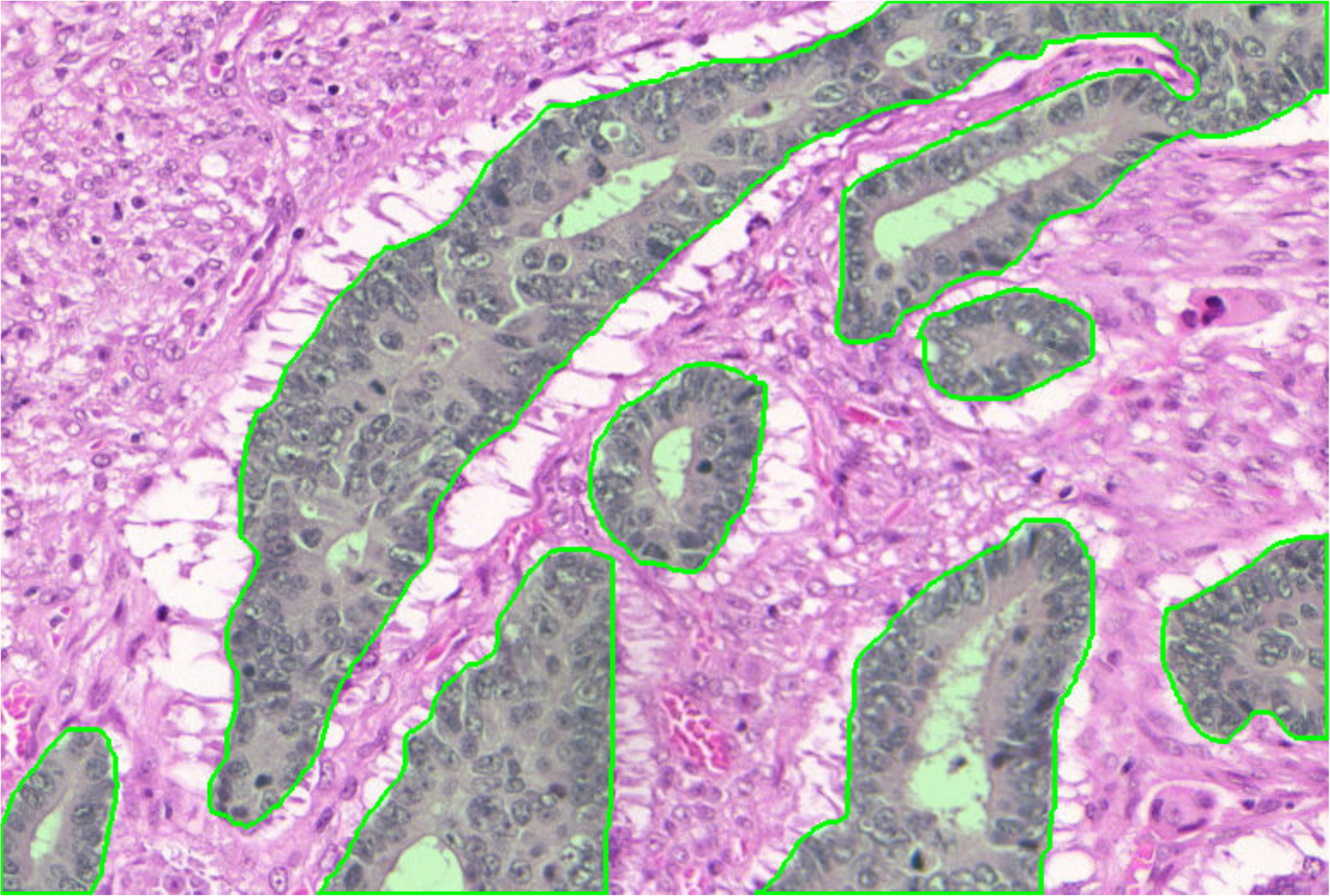} &
\includegraphics[width=0.32\linewidth]{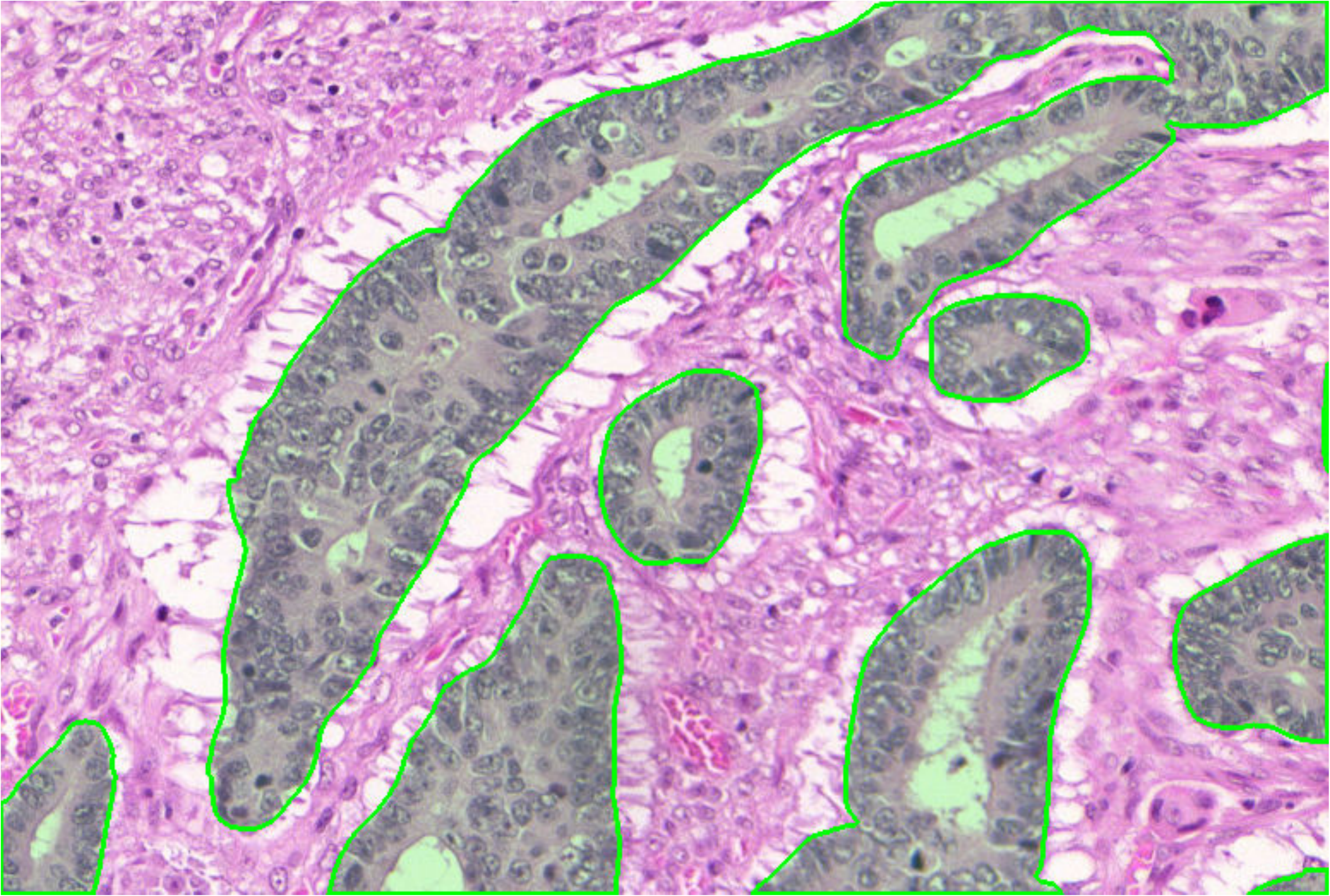} &
\includegraphics[width=0.32\linewidth]{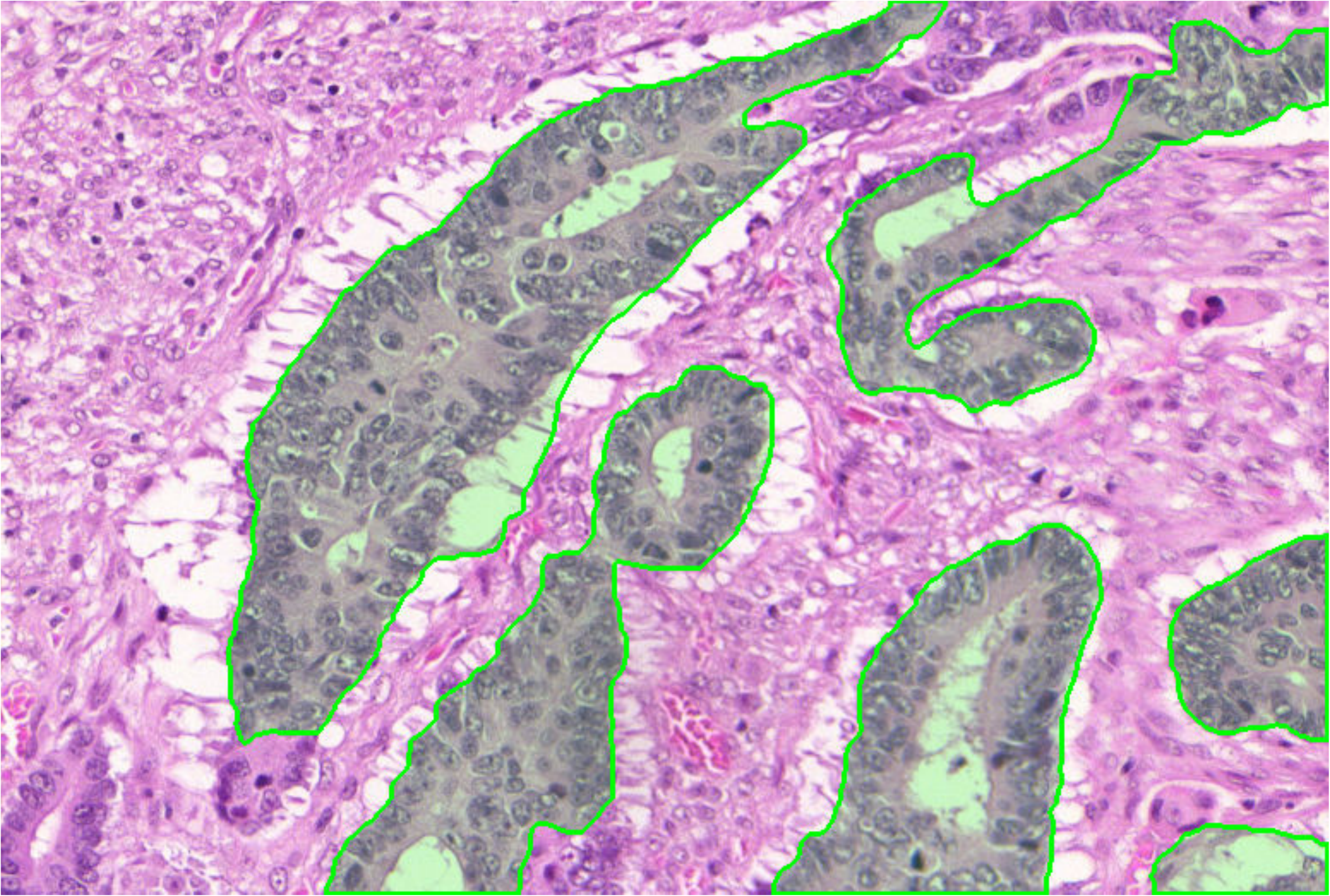} \\
& 
\renewcommand{\arraystretch}{0.8}
\renewcommand\tabcolsep{3pt}
\begin{tabular}{c}
{\footnotesize $\mathrm{F1score} = 0.875$} \\
{\footnotesize $\mathrm{Dice}_\mathrm{obj} = 0.961$} \\
{\footnotesize $\mathrm{H}_\mathrm{obj} = 11.480$} 
\end{tabular}
& 
\renewcommand{\arraystretch}{0.8}
\renewcommand\tabcolsep{3pt}
\begin{tabular}{c}
{\footnotesize $\mathrm{F1score} = 0.615$} \\
{\footnotesize $\mathrm{Dice}_\mathrm{obj} = 0.715$} \\
{\footnotesize $\mathrm{H}_\mathrm{obj} = 183.726$}
\end{tabular}
\end{tabular}



\caption{Example images showing segmentation results from some submitted entries. In each row, (left) ground truth, (middle) the best segmentation result, and (right) the worst segmentation result. For each image, the corresponding set of evaluation scores for the segmentation result is reported underneath the image.}
\label{fig:bestworst}
\end{figure}

Table \ref{tab:completeResults} summarizes the overall evaluation scores and ranks achieved by each entry from each test part. We list the entries according to the order of their rank sum, which indicates the overall performance across evaluation measures and tasks of the entries. The lower the rank sum, the more favorable the performance. The top three entries according to the overall rank sum in descending order are CUMedVision2, ExB1, and ExB3. However, if rank sum is considered with respect to the test part, the three best entries are  CUMedVision2, ExB2, and ExB3 for part A; whereas in part B, CUMedVision1, ExB1, and Freiburg2 come at the top. A summary of the ranking results from the competition can be found in \ref{sec:appendixA}. Some segmentation results and their corresponding evaluation scores are illustrated in Figure \ref{fig:bestworst} to give a better idea of how the evaluation scores correlate with the quality of the  segmentation.

\subsection{Additional Experiments} \label{sec:AdditionalExperiment}

\begin{figure}
\centering
\includegraphics[width=1\textwidth]{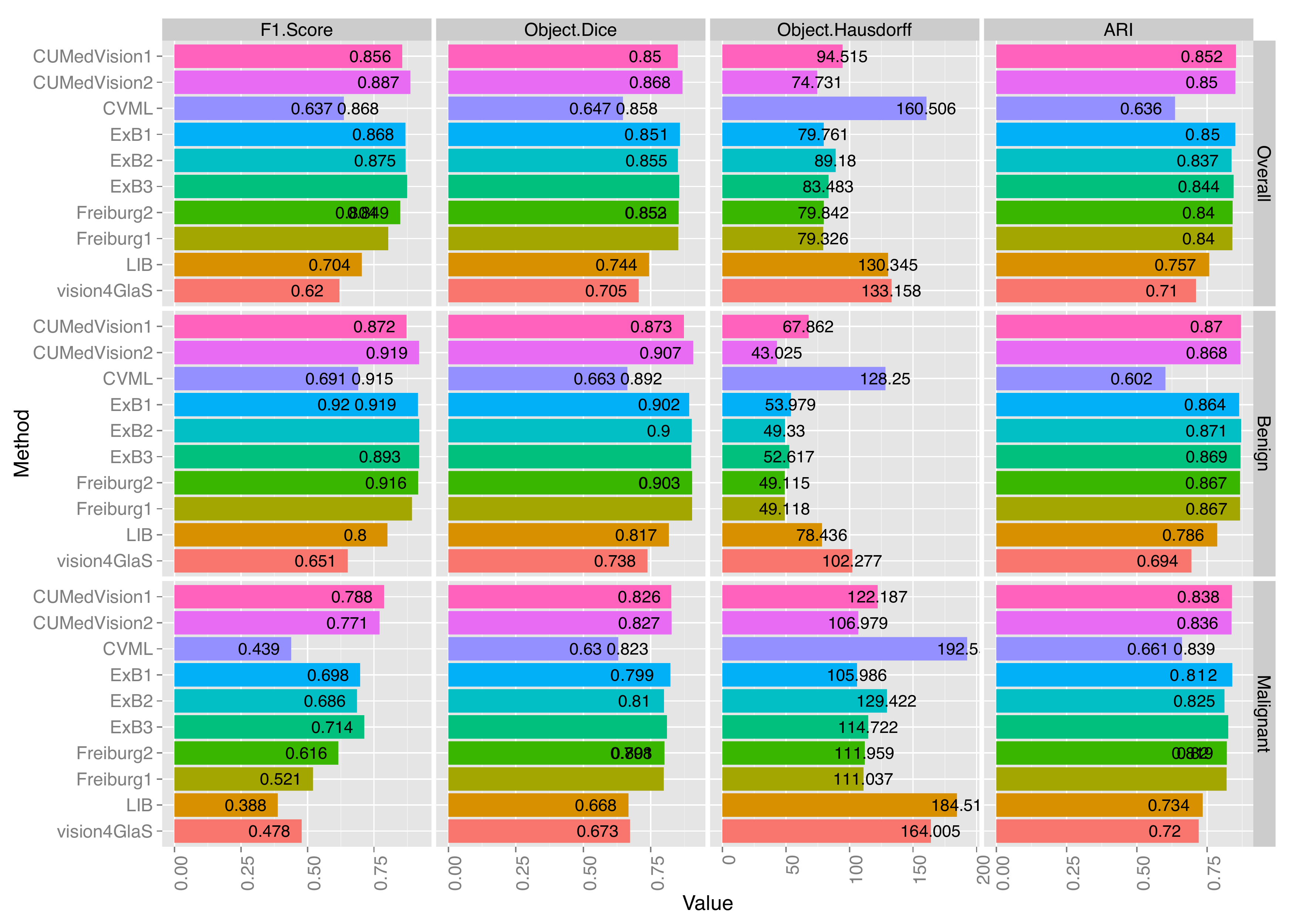}
\caption{Performance scores achieved by different entries on the combined test data. Evaluation is conducted on three subsets of the data: (1st row) the whole test data, (2nd row) benign, and (3rd row) malignant.}
\label{fig:results}
\end{figure}
\begin{table}[h]
\renewcommand{\arraystretch}{1.2}
\renewcommand\tabcolsep{3pt}
\centering
\caption{Ranking results of the entries when the two parts of test data are combined. Two set of ranking scheme are considered: a) $\mathrm{F1score} + \mathrm{Dice}_\mathrm{obj}$+$\mathrm{H}_\mathrm{obj}$ and b) $\mathrm{F1score} + \mathrm{ARI} + \mathrm{H}_\mathrm{obj}$. In addition to the evaluation on the whole test data (overall), the entries are evaluated on a subset of the data according to the histologic labels, i.e. benign and malignant.}
\label{tab:rank2}
\resizebox{0.9\columnwidth}{!}{
\begin{tabular}{|c|c|c|c|c|c|c|}
\hline
\multirow{3}{*}{Entry} & \multicolumn{6}{c|}{Final Ranking}                                                                            \\ \cline{2-7} 
                       & \multicolumn{3}{c|}{$\mathrm{F1score} + \mathrm{Dice}_\mathrm{obj}$+$\mathrm{H}_\mathrm{obj}$} & \multicolumn{3}{c|}{$\mathrm{F1score} + \mathrm{ARI} + \mathrm{H}_\mathrm{obj}$} \\ \cline{2-7} 
                       & Overall         & Benign         & Malignant         & Overall        & Benign        & Malignant       \\ \hline
CUMedVision1           & 7               & 7              & 3                 & 4              & 6             & 3               \\ \hline
CUMedVision2           & 1               & 1              & 1                 & 1              & 2             & 2               \\ \hline
CVML                   & 10              & 10             & 10                & 10             & 10            & 10              \\ \hline
ExB1                   & 2               & 6              & 2                 & 2              & 7             & 1               \\ \hline
ExB2                   & 6               & 3              & 7                 & 7              & 1             & 7               \\ \hline
ExB3                   & 3               & 5              & 4                 & 3              & 3             & 4               \\ \hline
Freiburg1              & 4               & 4              & 6                 & 6              & 5             & 6               \\ \hline
Freiburg2              & 5               & 2              & 5                 & 5              & 4             & 5               \\ \hline
LIB                    & 8               & 8              & 9                 & 8              & 8             & 9               \\ \hline
vision4GlaS            & 9               & 9              & 8                 & 9              & 9             & 8               \\ \hline
\end{tabular}
}
\end{table}

In the challenge, the split of the test data into two parts -- Part A (60 images) for off-site test and Part B (20 images) for on-site test -- to some extent introduces bias into the performance evaluation of the segmentation algorithms due to equal weight given to performance on the two test parts. The algorithms that perform particularly well on Test Part B would therefore get a better evaluation score even though they may not have performed as well on Test Part A, where the majority of the test dataset is to be found. In addition, the imbalance between the benign and malignant classes in Test Part B, only 4 benign (20\%) and 16 malignant (80\%) images, would also favor algorithms that perform well on the malignant class. In order to alleviate these issues, we merged the two test parts and re-evaluated the performance of all the entries. In addition, as suggested by one of the participating teams, the adjusted Rand index is included as another performance measurement for segmentation.

The evaluation scores calculated from the combined two test parts are presented as bar chart in Figure \ref{fig:results}. The final rankings based on the rank sums of evaluation scores calculated from the combined two test parts are reported in Table \ref{tab:rank2}. Here, two set of rank sums are considered: one calculated according to the criteria of the competition (i.e., $\mathrm{F1score} + \mathrm{Dice}_\mathrm{obj} + \mathrm{H}_\mathrm{obj}$), and the other where the adjusted Rand index is used instead of the object-level Dice index to evaluate segmentation accuracy (i.e., $\mathrm{F1score} + \mathrm{ARI} + \mathrm{H}_\mathrm{obj}$). For both sets of rank sums, the new ranking orders are largely similar to those reported in Section \ref{sec:EvaluationResults}, with a few swaps in the order, while the top three entries remaining the same, namely CUMedVision2, ExB1, ExB3. 

\begin{figure}
\begin{subfigure}{\textwidth}
\centering
\includegraphics[height=0.28\linewidth]{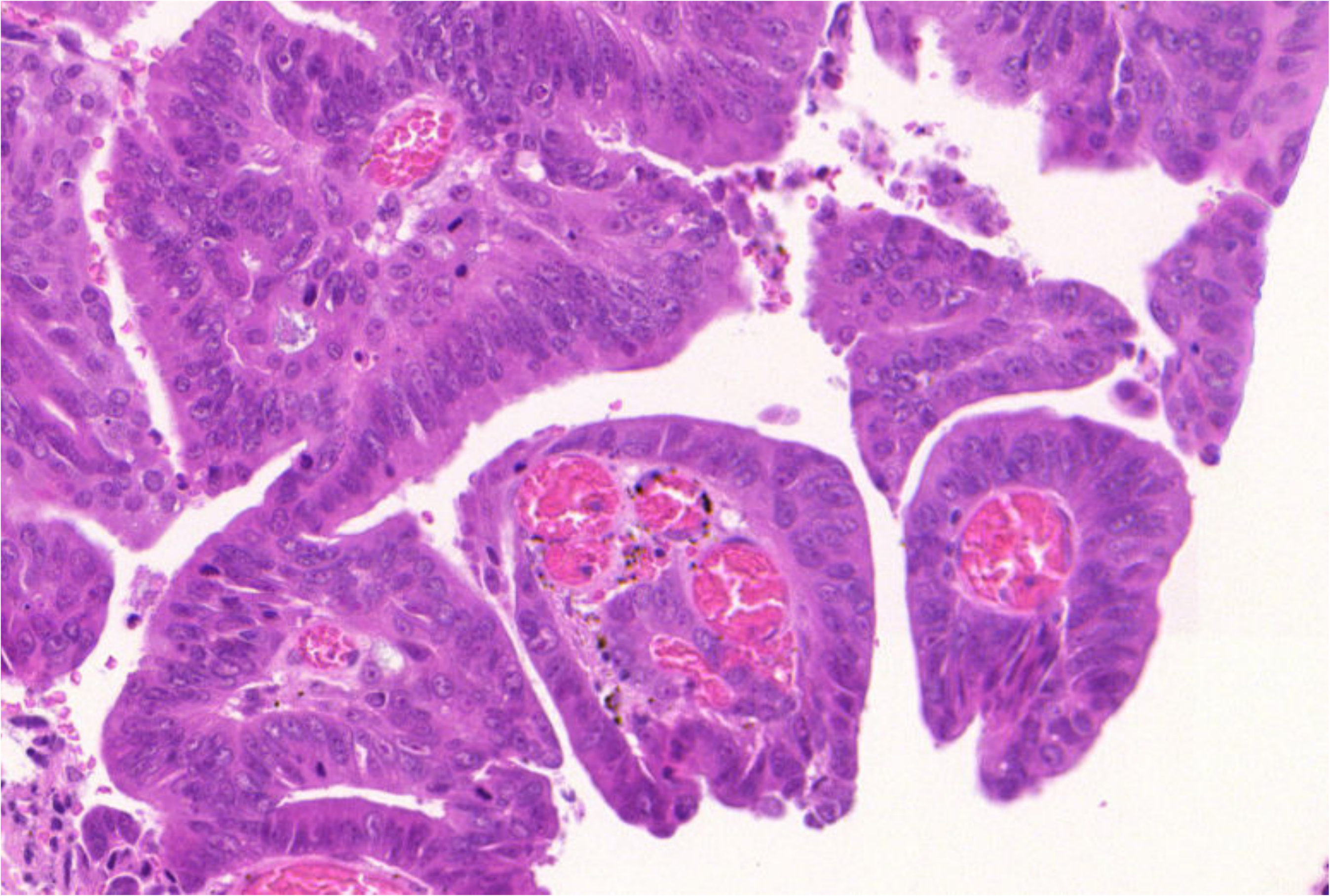}
\includegraphics[height=0.28\linewidth]{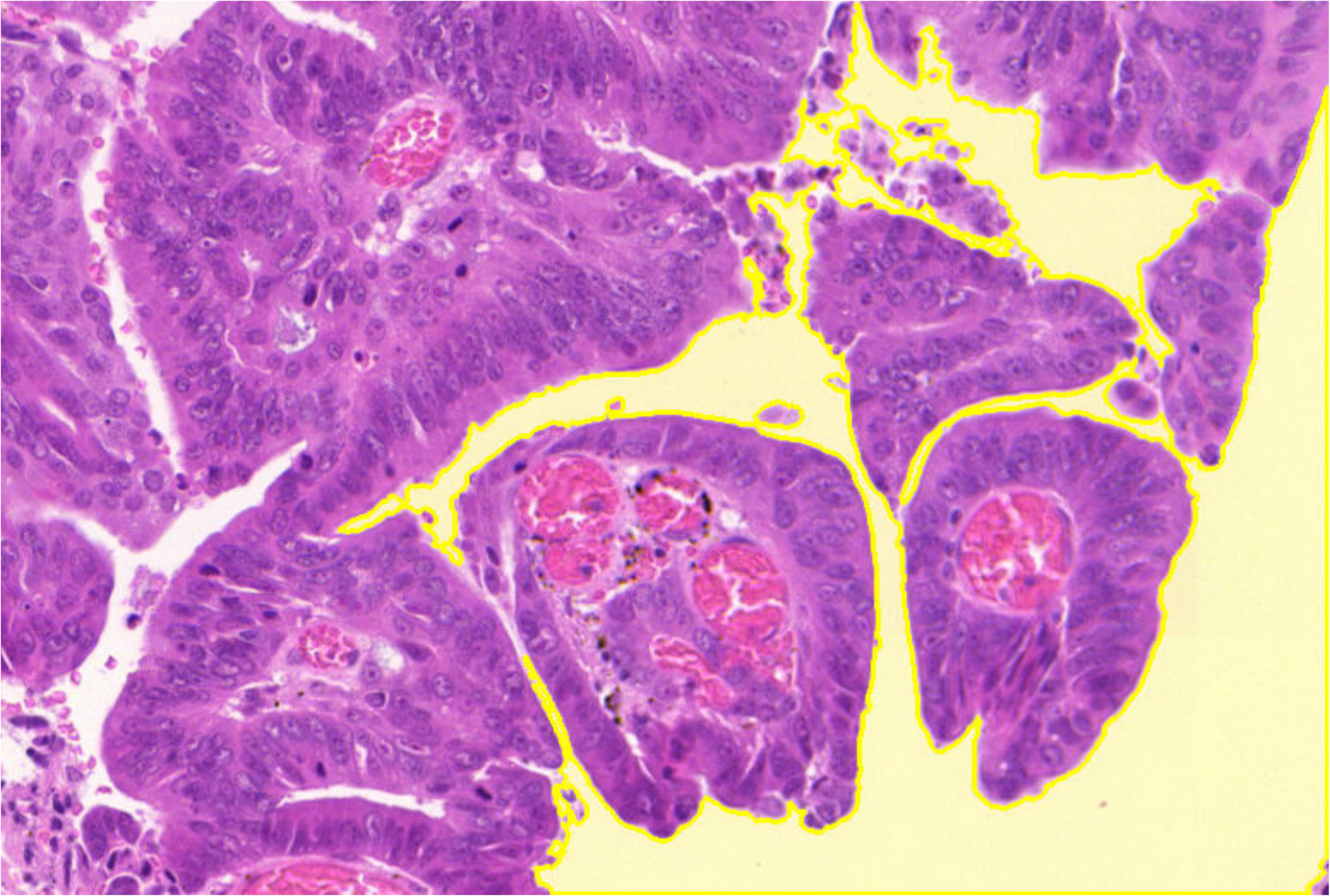}
\caption{}
\label{fig:empty}
\end{subfigure}

\begin{subfigure}{\textwidth}
\centering
\includegraphics[height=0.28\linewidth]{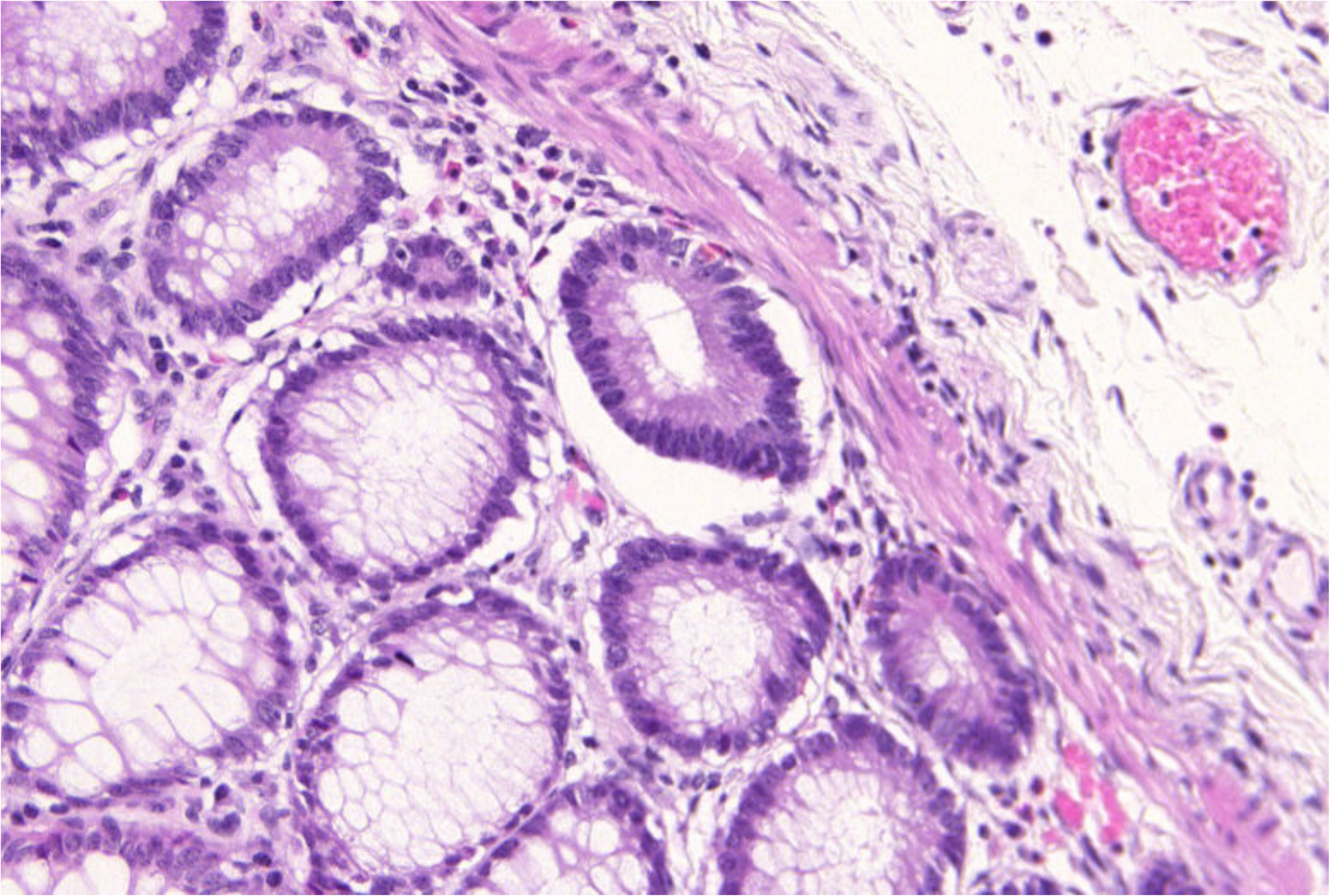}
\includegraphics[height=0.28\linewidth]{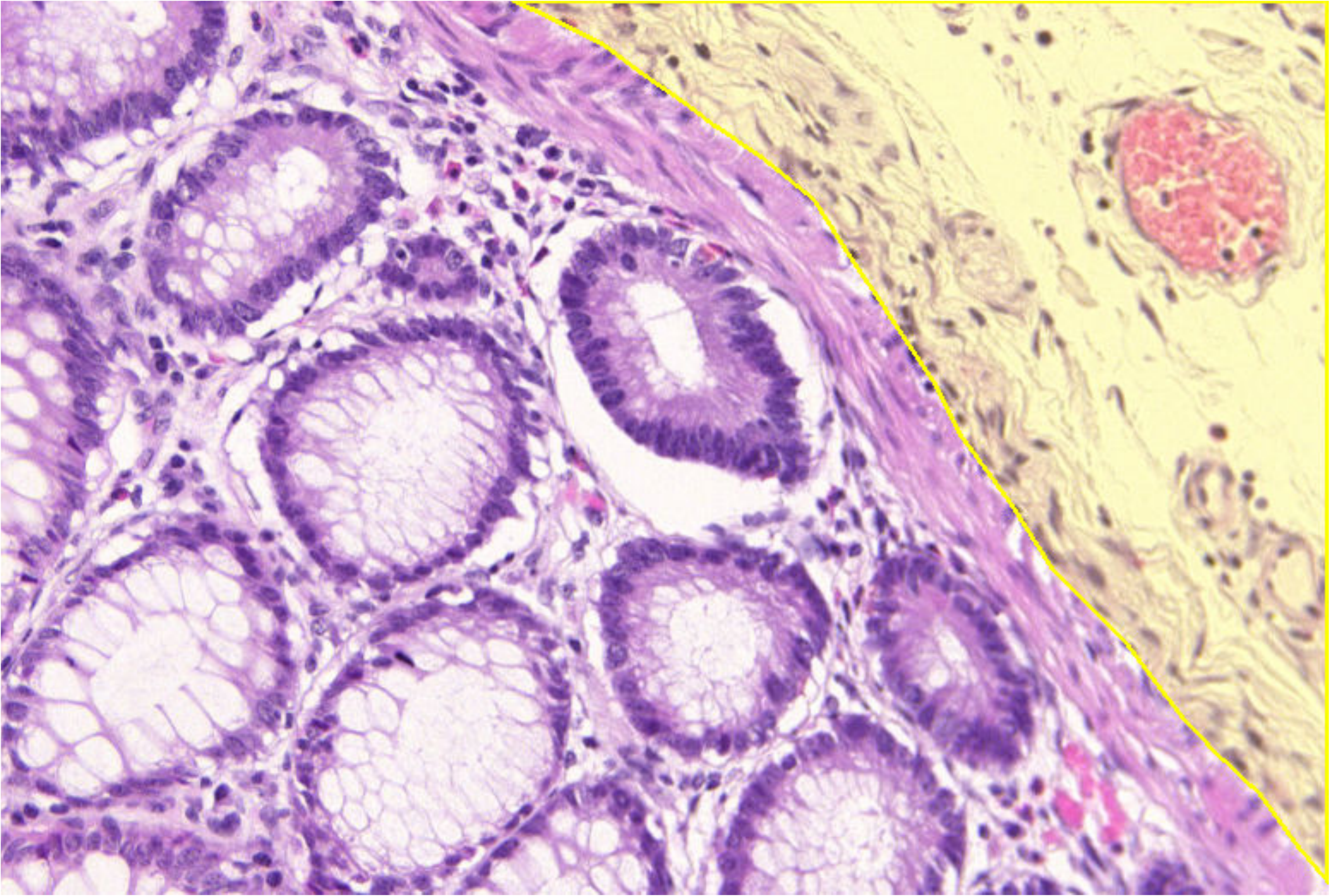}
\caption{}
\label{fig:mucosa}
\end{subfigure}

\begin{subfigure}{\textwidth}
\centering
\includegraphics[height=0.28\linewidth]{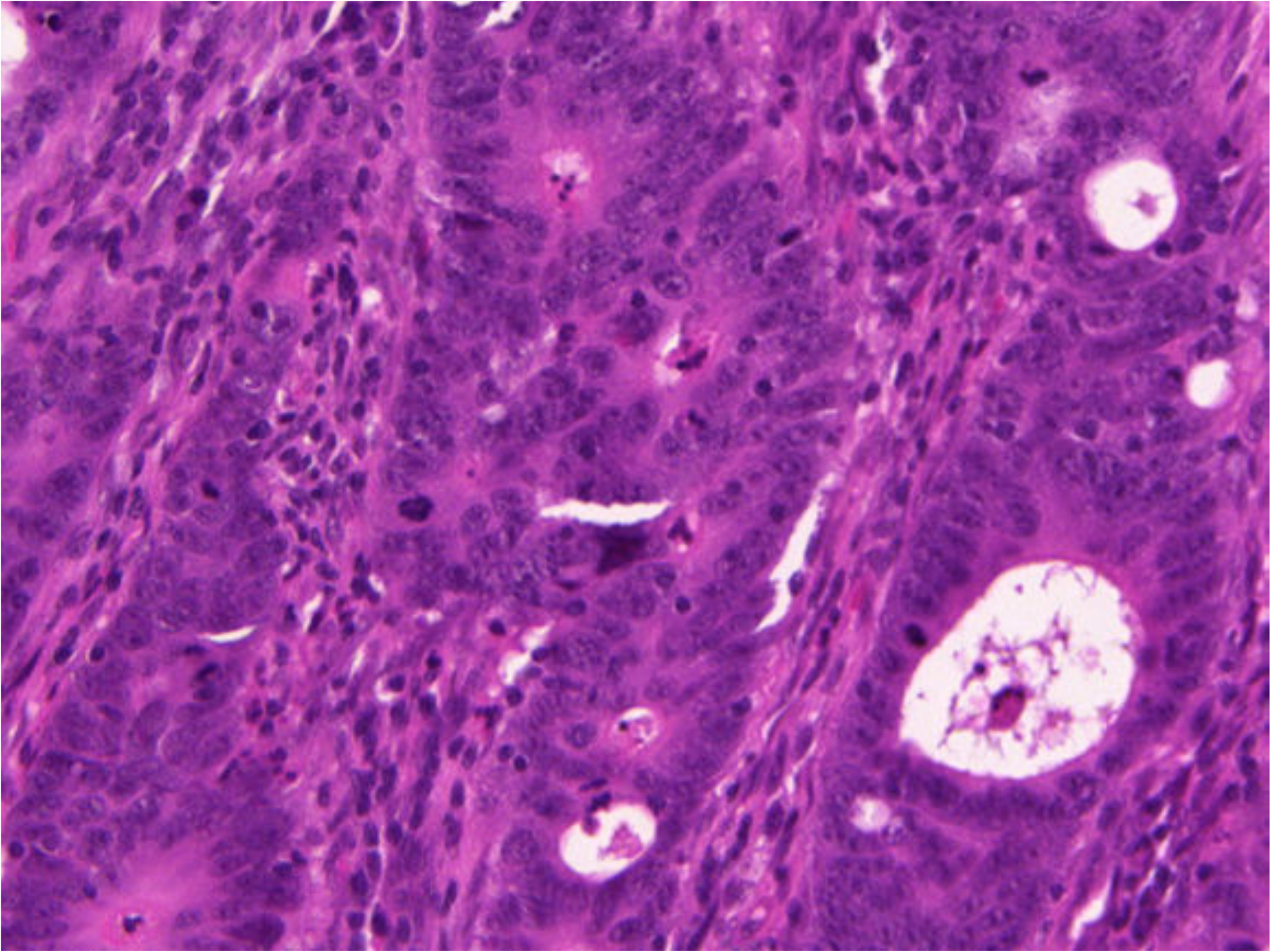}
\includegraphics[height=0.28\linewidth]{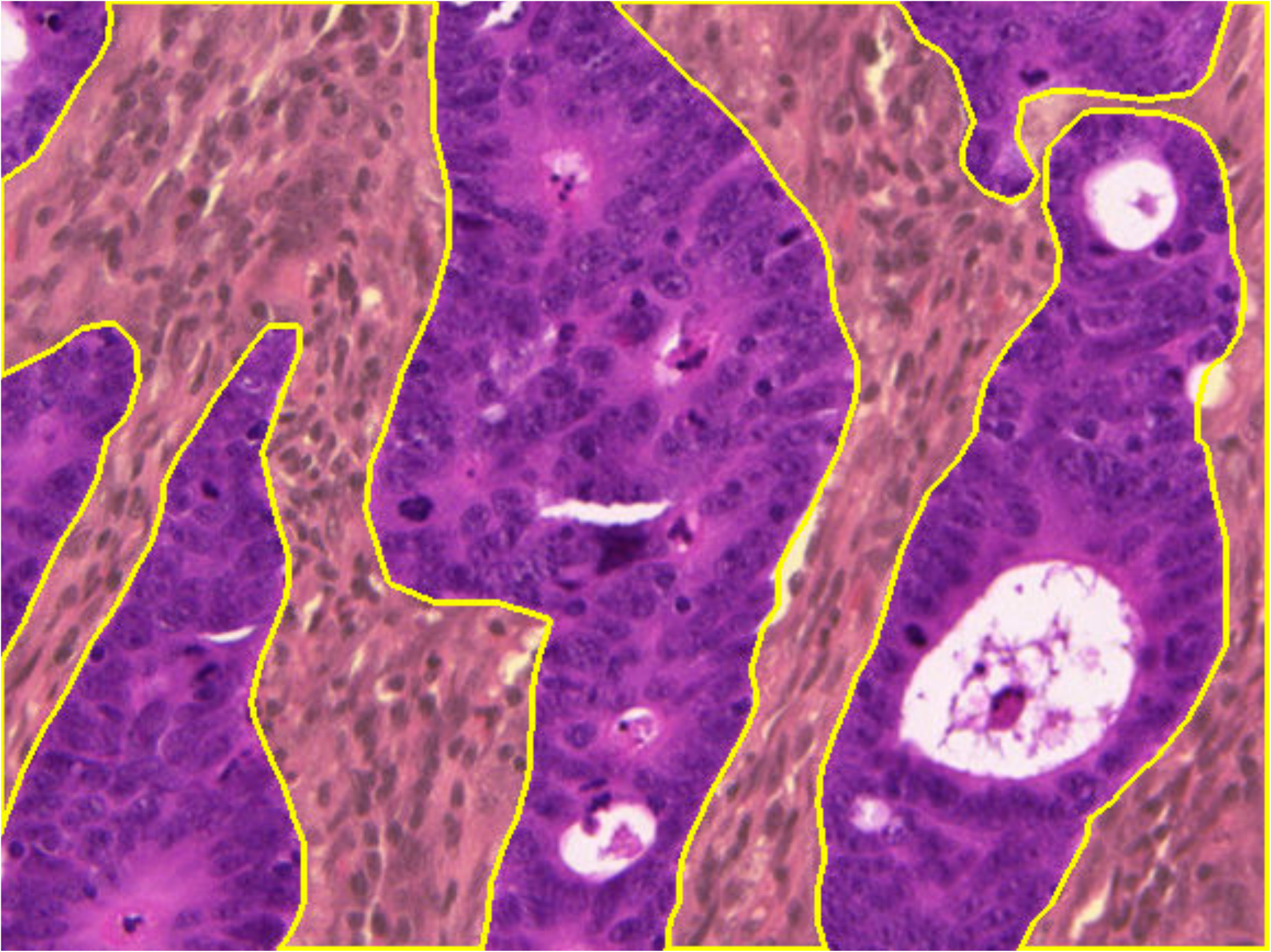}
\caption{}
\label{fig:fgbg}
\end{subfigure}  

\begin{subfigure}{\textwidth}
\centering
\includegraphics[height=0.28\linewidth]{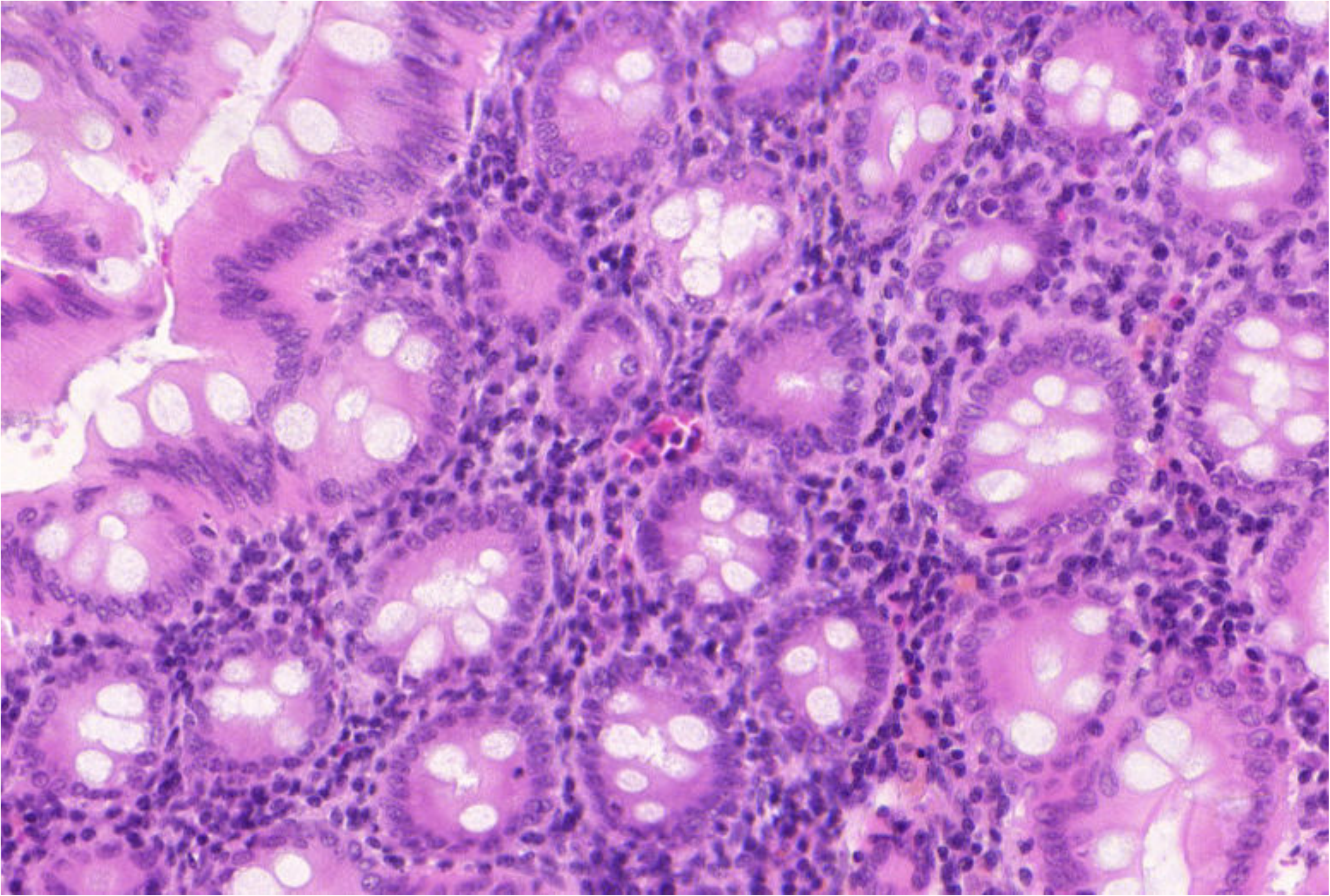}
\includegraphics[height=0.28\linewidth]{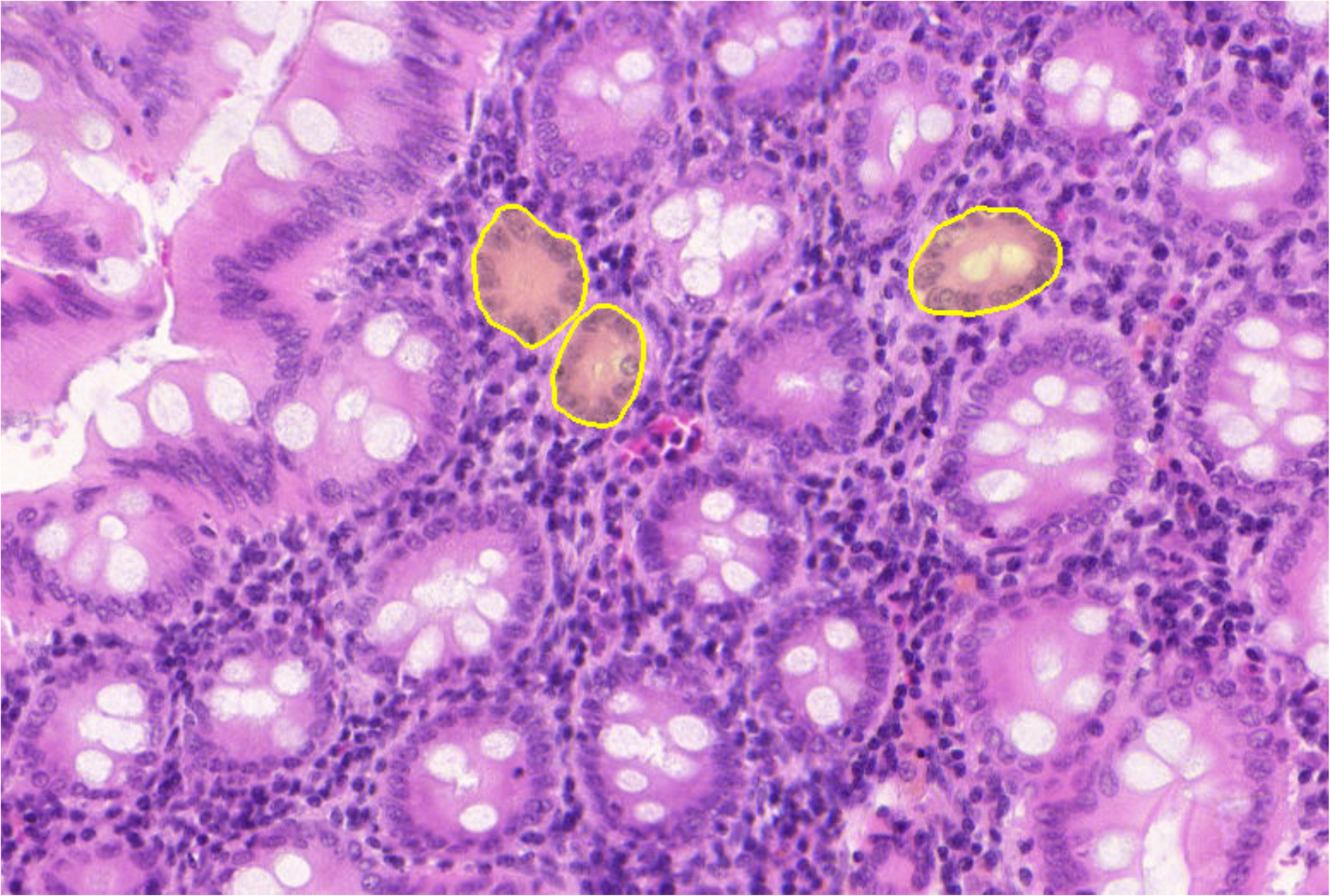}
\caption{}
\label{fig:area}
\end{subfigure}  

\caption{Example images showing some challenging features in the dataset: (a) lumen of the gastrointestinal tract, (b) sub-mucosa layer, (c) area with dense nuclei in mucosa layer, and (d) small glands. Each example is shown with (left) the original image and (right) the overlaid image highlighting the area with challenging characteristic.} \label{fig:problem}
\end{figure}

The main factors that negatively affect the performance of the methods are a number of challenges presented by the dataset. Firstly, large white empty areas corresponding to the lumen of the gastrointestinal tract which are not in the interior of intestinal glands can easily confuse the segmentation algorithms (Figure \ref{fig:empty}). Secondly, characteristics of non-glandular tissue can sometimes resemble that of the glandular tissue. For instance, connective tissue in muscularis mucosa or sub-mucosa layers of the colon is stained white and pinkish and has less dense nuclei, thus resembling the inner part of glands (Figure \ref{fig:mucosa}). In the case where there is less stain contrast between nuclei and cytoplasm due to elevated levels of Hematoxylin stain, non-glandular tissue with dense nuclei can look similar to malignant epithelial tissue (Figure \ref{fig:fgbg}). Thirdly, small glandular objects are blended into the surrounding tissue and can be easily mis-detected (Figure \ref{fig:area}). A careful inspection of the segmentation results generated by each entry showed that methods by CUMedVision, ExB, and Freiburg better avoid over-segmentation or under-segmentation when facing the above-mentioned pitfalls.

The performance of each entry with respect to the histologic grade of cancer was also examined. Their evaluation scores based on benign and malignant samples are reported in the second and the third rows of Figure \ref{fig:results} respectively, and the ranking orders derived from the rank sums of the scores are shown in Table \ref{tab:rank2}. 
Based on these results, one can get a better contrast between the performance of the entries that enforce border separation and those that do not. By applying a predicted border mask to separate clumped segmented objects, CUMedVision2 performs better than CUMedVision1, which tends to produce segmentation results that merge neighboring glands together, in both benign and malignant cases. Similarly, ExB1 is able to segment malignant glands better than ExB2 and ExB3 that do not utilize border separation. However, this can have an adverse effect if the algorithm already yields segmentation results that separate individual objects well, such as in the case of ExB1 which under-segments benign glandular objects as compared to its counterparts ExB2 and ExB3.


\subsection{General Discussion}
The objectives of this challenge were to raise the research community's awareness of the existence of the intestinal gland segmentation problem in routine stained histology images, and at the same time to provide a platform for a standardized comparison of the performance of automatic and semi-automatic algorithms. The challenge attracted a lot of attention from researchers, as can be seen from the number of registered teams/individuals and the number of submissions at each stage of the competition. Interestingly, some of the teams had no experience in working with histology images before. We would like to emphasize that finding the best performing approach is not the main objective of the competition, but rather pushing the boundaries of the-state-of-the-art approaches. Already, we have seen quite interesting developments from many participating teams and the leading algorithms have produced excellent results, both qualitatively and quantitatively.

As noted in the Introduction, morphometric analysis of the appearance of cells and tissues, especially those forming glands from which tumors  originate, is one of the key components towards precision medicine, and segmentation is the first step to attain morphological information. Some may have argued that there is no need to perform segmentation, but instead, to follow conventional pattern recognition approaches by extracting mathematical features which normally capture local and/or global tissue architecture and then identifying features that are most suited to the objective of the study. It is true that there are a number of successful works that follow such an approach \citep{jafari2003multiwavelet,tabesh2007multifeature,altunbay2010color,basavanhally2010computerized,ozdemir2013hybrid,gultekin2015two}. However, because these extracted features are often physically less interpretable in the eyes of practitioners, it is difficult to adopt such an approach in clinical settings. On the other hand, the appearance of glands such as size and shape obtained through segmentation is easy to interpret. Segmentation also helps to localize other type of information (e.g., texture, spatial arrangement of cells) that is specific to the glandular areas.

Even though the dataset used in the challenge included images of different histologic grades taken from several patients, it lacked other aspects. First of all, inter-observer variability was not taken into account as the ground truth was generated by a single expert. This is because the intricate and arduous nature of the problem makes it difficult to find several volunteer experts to perform manual segmentation. Considerable experience is required in order to delineate boundaries of malignant glands, which are not so well-defined as those of the benign ones. Moreover, a single image can contain a large number of glands to be segmented, making the task very laborious. Secondly, digitization variability was also not considered in this dataset. It is, in fact, very important to evaluate the robustness of algorithms when the data are scanned by different instruments. As whole-slide scanners are becoming increasingly available, this type of real-world problem should be expected.

The choice of evaluation measures would also affect the comparative results. In this challenge, we emphasized object segmentation and accordingly defined the object-level Dice index and the object-level Hausdorff distance to measure segmentation accuracy at the object level rather than at the pixel level. Nonetheless, it has been suggested that these measures are too strict, as they put a severe penalty on mismatch of the objects. One could replace these measures by less conservative ones, for example, adjusted Rand index \citep{hubert1985comparing} or a topology preserving warping error \citep{jain2010boundary} for a volume-based metric and elastic distance \citep{younes1998computable,joshi2007removing} for a boundary-based metric. For this reason, we included adjusted rand index as an alternative to object-level Dice index in Section \ref{sec:AdditionalExperiment}. As we have already pointed out, this results in only a minor change in the ranking order of the entries. Another aspect that was not explicitly included in the evaluation was execution times. Nevertheless, all the algorithms were capable of completing the segmentation task on the on-site test data (Part B) in the given amount of time with or without limitation of resources. Time efficiency is required to process large scale data, such as whole-slide images, whose volume is growing by the day as slides are routinely scanned. Still, in medical practice, accuracy is far more important than speed.

It is worth noting that the used evaluation metrics used here are clinically relevant. As mentioned in the Introduction, morphology of intestinal glands is the key criterion for colorectal cancer grading. This includes shape, size, and  formation of the glands. Thus, in terms of clinical relevance, the object-Hausdorff distance is used in accessing the shape similarity between the segmentation results and the ground truth. The object-Dice index is used in assessing the closeness between the volume of the segmentation results and that of the ground truth, which is important in estimating the size of individual glands. Although not directly clinically relevant, F1 score is important in assessing the accuracy of the identified glands. Since the morphological assessment is done on the basis of tissue slide including several thousands of glands, an algorithm with high value of F1 score is more preferable as it can detect a larger number of glands. 

Gland segmentation algorithms presented here are not ready for deployment into clinic in their present form. Although some of the top algorithms produce good segmentation results for the contest dataset and will probably fare well in the real world, there needs to be a large-scale validation involving data from multiple centers annotated by multiple pathologists before any of these algorithms can be deployed in a diagnostic application.

The challenge is now completed, but the dataset will remain available for research purposes so as to continually attract newcomers to the problem and to encourage development of state-of-the-art methods. Extension of the dataset to address inter-observer and inter-scanner variability seems to be the most achievable aim in the near future. Beyond the scope of segmentation, there lie various extremely interesting future research directions. Previous studies have shown the strong association between the survival of colorectal cancer patients and tumor-related characteristics, including lymphocytic infiltration \citep{galon2006type,fridman2012immune}, desmoplasia \citep{tommelein2015cancer}, tumor budding \citep{mitrovic2012tumor}, and necrosis \citep{richards2012prognostic}. A systematic analysis of these characteristics with the help of gland segmentation as part of automatic image analysis framework could lead to a better understanding of the relevant cancer biology as well as bring precision and accuracy into assessment and prediction of the outcome of the cancer.
\section{Conclusions}
This paper presented a summary of the Gland Segmentation in Colon Histology Images (GlaS) Challenge Contest which was held in conjunction with the 18th International Conference on Medical Image Computing and Computer Assisted Interventions (MICCAI'2015). The goal of the challenge was to bring together researchers interested in the gland segmentation problem, to validate the performance of their existing or newly invented algorithms on the same standard dataset. In the final round, the total number of submitted entries for evaluation was 19, and we presented here in this paper 10 of the leading entries. The dataset used in the challenge has been made publicly available and can be accessed at the challenge website (\url{http://www.warwick.ac.uk/bialab/GlasContest/}). Those who are interested in developing or improving their own approaches are encouraged to use this dataset for quantitative evaluation.
\section{Acknowledgements}

This paper was made possible by NPRP grant number NPRP5-1345-1-228 from the Qatar National Research Fund (a member of Qatar Foundation). The statements made herein are solely the responsibility of the authors. Korsuk Sirinukunwattana acknowledges the partial financial support provided by the Department of Computer Science, University of Warwick, UK. CUMedVision team acknowledges Hong Kong RGC General Research Fund (Project No. CUHK 412513). The work of Olaf Ronneberger was supported by the Excellence Initiative of the German Federal and State Governments (EXC 294). Philipp Kainz was supported by the Excellence Grant 2014 of the Federation of Austrian Industries (IV), and Martin Urschler acknowledges funding by the Austrian Science Fund (FWF): P 28078-N33.

The authors thank Professor David Epstein for his extensive help with the wording and mathematical notation in this paper, and Nicholas Trahearn for his help with the wording.

\appendix
\section{The Complete Contest Results} \label{sec:appendixA}
\begin{figure}[h]
\centering
\includegraphics[width=1\textwidth]{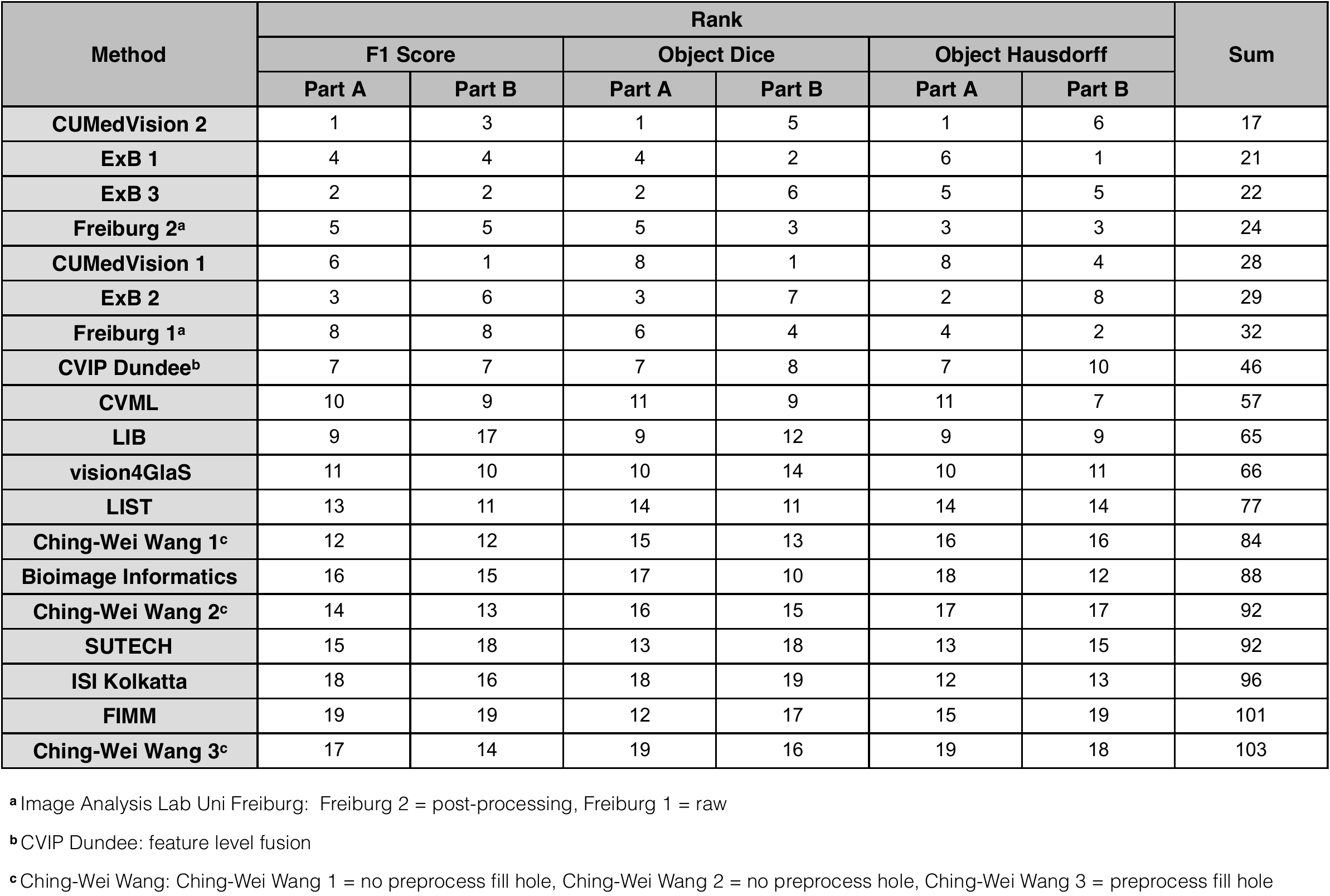}
\caption{The ranking results from the GlaS Challenge Contest.}
\label{fig:glas2}
\end{figure}

A summary of the ranking results from the contest is given in Figure \ref{fig:glas2}.

\bibliographystyle{plainnat}
\bibliography{main_preprint_arxiv_2016_08_30.bib}

\end{document}